\pdfoutput=1

\documentclass[11pt]{article}

\usepackage[final]{acl}

\usepackage{times}
\usepackage{latexsym}
\usepackage{multirow}
\usepackage{booktabs}
\usepackage{amsmath}
\usepackage{amssymb}
\usepackage{adjustbox}
\usepackage{amsfonts}

\usepackage{cleveref}
\usepackage{graphicx}
\usepackage{epstopdf}
\usepackage{hyperref}
\usepackage{colortbl}
\usepackage{float}
\usepackage{pifont}
\usepackage{xcolor}
\definecolor{my_green}{RGB}{51,102,0}
\definecolor{my_red}{RGB}{204, 0, 0}
\renewcommand{\checkmark}{\textcolor{my_green}{\ding{51}}} %
\newcommand{\crossmark}{\textcolor{my_red}{\ding{55}}} %

\usepackage[T1]{fontenc}

\usepackage[utf8]{inputenc}

\usepackage{microtype}

\usepackage{inconsolata}

\usepackage{graphicx}

\title{MegaPairs: Massive Data Synthesis For Universal Multimodal Retrieval} 

\author{Junjie Zhou$^1$, \ \ \ Zheng Liu$^{2,5}$\thanks{Corresponding author.}, \ \ \ Ze Liu$^{3}$, \ \ \ \textbf{Shitao Xiao}$^2$, \ \ \ Yueze Wang$^{2}$, \ \ \ 
 \textbf{Bo Zhao}$^{2,4}$ \\
\textbf{Chen Jason Zhang}$^{5}$, \ \ \ \textbf{Defu Lian}$^{3}$, \ \ \ \textbf{Yongping Xiong}$^1$  \\
        $^1$ Beijing University of Posts and Telecommunications, \ $^2$ Beijing Academy of Artificial Intelligence, \\ $^3$ University of Science and Technology of China, \ \ $^4$ Shanghai Jiaotong University \\
        $^5$ The Hong Kong Polytechnic University \\
        \texttt{zhoujunjie@bupt.edu.cn} \quad
        \texttt{zhengliu1026@gmail.com} 
}

\begin{document}
\maketitle
\begin{abstract}

Despite the rapidly growing demand for multimodal retrieval, progress in this field remains severely constrained by a lack of training data. In this paper, we introduce \textbf{MegaPairs}, a novel data synthesis method that leverages vision language models (VLMs) and open-domain images, together with a massive synthetic dataset generated from this method. Our empirical analysis shows that MegaPairs generates high-quality data, enabling the multimodal retriever to significantly outperform the baseline model trained on 70$\times$ more data from existing datasets. Moreover, since MegaPairs solely relies on general image corpora and open-source VLMs, it can be easily scaled up, enabling continuous improvements in retrieval performance. In this stage, we produced more than 26 million training instances and trained several models of varying sizes using this data. These new models achieve state-of-the-art zero-shot performance across 4 popular composed image retrieval (\textbf{CIR}) benchmarks and the highest overall performance on the 36 datasets provided by \textbf{MMEB}. They also demonstrate notable performance improvements with additional downstream fine-tuning. Our produced dataset, well-trained models, and data synthesis pipeline will be made publicly available to facilitate the future development of this field.  
\end{abstract} 


\section{Introduction}


Multimodal retrieval is a critical research problem for IR and AI communities. It aims to satisfy people's information needs across different data modalities, especially texts and images. Nowadays, multimodal retrieval has been applied to a wide variety of real-world scenarios, such as image search~\cite{mscoco-chen2015microsoft,fashioniq-wu2021fashion,magiclens}, visual question answering (VQA)~\cite{ok-vqa,mathew2021docvqa}, and retrieval-augmented generation (RAG) of vision language models~\cite{murag-DBLP:conf/emnlp/ChenHCVC22, yu2024visrag}. Given the widespread application scenarios, it's necessary to develop universal multimodal retrievers which can uniformly support any task requirements and working domains. 

The progress of universal multimodal retrievers have been substantially advanced on top of the pre-trained vision-languages models, like CLIP~\cite{clip-radford2021learning}, ALIGN~\cite{align2021}, and SigLIP~\cite{siglip2023}. These models are pre-trained to produce discriminative and unified representations for texts and images, thus creating a solid foundation for multimodal retrieval. However, the existing vision-language encoders are mostly pre-trained from text-image matching tasks. Although these models have achieved an initial capability for text-to-image retrieval~\cite{flickr-young2014image,mscoco-chen2015microsoft}, they are insufficient for other common multimodal tasks, such as composed image retrieval~\cite{cirr-liu2021image,circo,magiclens} and multimodal document retrieval~\cite{chang2022webqa,univl-liu2022universal}. 

To enhance the multi-task capacity, fine-tuning pre-trained models with comprehensive instructions, commonly known as instruction-tuning, has gained significant popularity. This approach was first applied in the supervised fine-tuning of large language models (LLMs) \cite{ouyang2022training,wei2021finetuned,chung2024scaling}, and later introduced for training text embeddings \cite{su2022one,asai2022task,zhang2023retrieve,bge}. Building on these successes, instruction-tuning has been further extended to multimodal embedding models \cite{mbeir-wei2023uniir,sharifymoghaddam2024unirag}, where pre-trained vision-language encoders are continually fine-tuned using a variety of multimodal retrieval instructions. Given the scarcity of instruction-tuning data for embedding models, researchers have proposed leveraging LLMs to generate synthetic data from Internet resources \cite{e5-mistral}. In the field of multimodal retrieval, a notable example is presented by MagicLens \cite{magiclens}, which synthesizes open-ended search instructions for co-existing images within the same webpage. 


Despite recent advancements by MagicLens, current data synthesis methods still face significant limitations in data \textit{scalability}, \textit{quality}, \textit{diversity}, and \textit{availability}. Specifically, only a small fraction of webpages on the internet contain multiple images (scalability), not to mention that many of these co-existing images are either unrelated or near-duplicates (quality). Besides, the remaining correlated images often exhibit monotonous relationships, such as different angles of the same object (diversity). Finally, large-scale instruction-tuning datasets for multimodal retrieval are typically held privately by individual research labs (availability). 

In this paper, we introduce a novel data synthesis method called \textbf{MegaPairs}, accompanied by a large-scale instruction dataset generated using this approach. MegaPairs is distinguished by its construction of \textit{a heterogeneous KNN triplet for open-domain images}. Particularly, it leverages three different similarity models to sample correlated image pairs, including CLIP vision-encoder for visual-semantic correlations~\cite{evaclip-sun2023eva}, DINO vision-encoder for visual-pattern correlations~\cite{dinov2}, and CLIP text-encoder for caption correlations. The sampled image pairs are presented for the VLM and LLM annotators, which generate comprehensive descriptions of the relationships between the two images and create pseudo-retrieval instructions based on the descriptions. This approach enables a huge amount of instructions to be generated for a general dataset, like Datacomp~\cite{datacomp}, which significantly improves the scalability of data synthesis. It also introduces diverse instructions of guaranteed quality, given its sampling of heterogeneous relationships from open-ended image corpora. Additionally, by utilizing open-source VLM and LLM models (e.g., InternVL2-26B~\cite{internvl2}, Llama-3-8B~\cite{llama3}), the entire process can operate at a low cost. 


We've produced 26 million data instances in this stage, achieving superior data quality compared to the existing datasets. In our pilot experiment, with just 500K sampled instances from MegaPairs, the same pre-trained model's fine-tuning performance already surpasses that of the entire 36.7M training instances from MagicLens, i.e., delivering better results with 70$\times$ less training data. We further trained three multimodal retrievers, \textbf{MMRet}, of varying sizes based on the whole synthetic dataset and perform comprehensive evaluations with a wide range of multimodal retrieval tasks. Remarkably, MMRet achieved state-of-the-art performance on 4 popular composed image retrieval (CIR) benchmarks and the 36 datasets provided by MMEB \cite{vlm2vec2024} in the zero-shot setting. Furthermore, the models demonstrated substantial improvements and maintain leading positions after downstream fine-tuning. The entire suite of assets, including the dataset, the well-trained models, and the data production pipeline, will be made publicly available to advance the future progress in this field.

\section{Related Work}
\paragraph{Multimodal Retrieval.}

Traditionally, retrieval tasks have focused on scenarios where queries and candidates exist in distinct modalities, such as unimodal retrieval~\cite{thakur2021beir} and cross-modal retrieval~\cite{mscoco-chen2015microsoft}. However, there is a growing demand for multimodal retrieval tasks, where queries or candidates integrate both image and text modalities. These tasks have wide applications, including image retrieval with instructions \cite{fashioniq-wu2021fashion, cirr-liu2021image, magiclens}, multimodal document retrieval~\cite{chang2022webqa, univl-liu2022universal}, knowledge retrieval with multimodal queries~\cite{remuq-DBLP:conf/acl/0003FGYB23}, and retrieval-augmented generation \cite{racm3-DBLP:conf/icml/YasunagaAS0LLLZ23, yu2024visrag}. Most existing methods employ pre-trained vision-language models (VLMs) to address these tasks~\cite{clip-radford2021learning,blip2-li2023blip,saito2023pic2word}. However, the common VLMs are purely trained on image-text matching datasets~\cite{cc12m-changpinyo2021conceptual,laion5b-DBLP:conf/nips/SchuhmannBVGWCC22}, which are in lack of ability to jointly encode and comprehend both modalities effectively. As a result, it is necessary to create proper datasets so as to extend VLMs for the diversified multimodal retrieval tasks.

\begin{figure*}[tb]
    \centering
    \includegraphics[width=\textwidth]{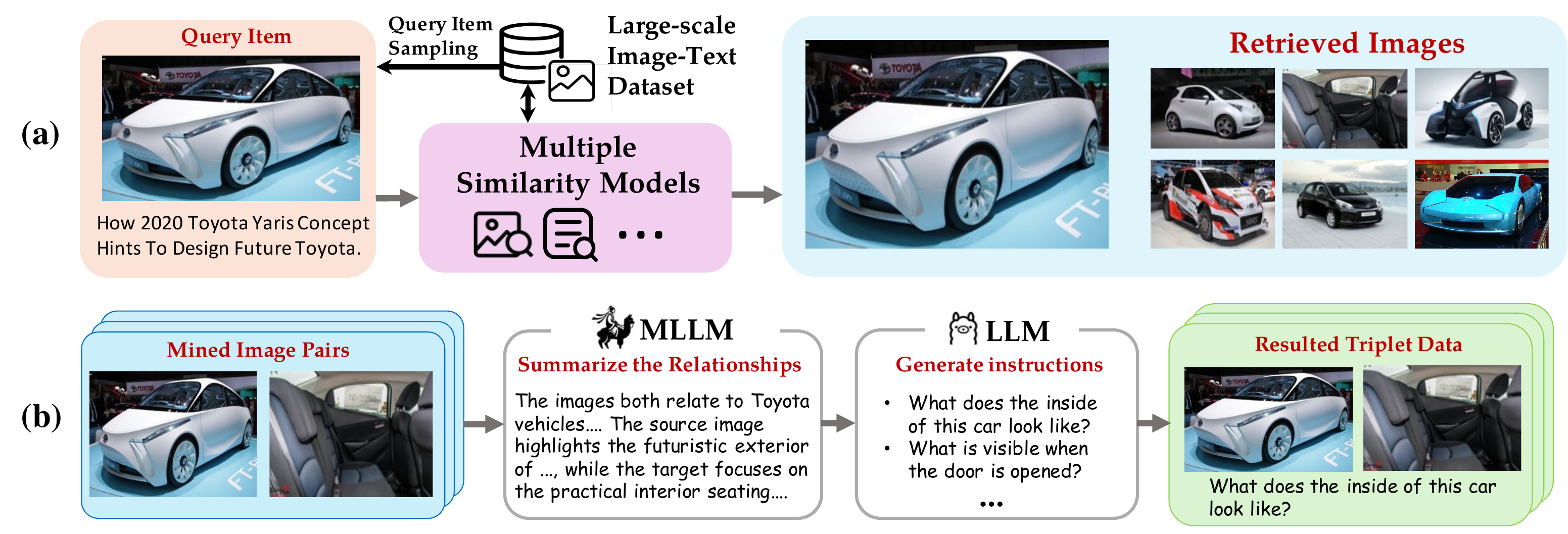}
    \vspace{-0.6cm}
    \caption{Construction pipeline of multimodal triplets: \textbf{(a)} mining of image pairs, \textbf{(b)} generation of open-ended instructions. Multiple similarity models are used to introduce diversified correlations for the image pairs.} 
    \vspace{-0.6cm}
    \label{fig:pipeline}
\end{figure*} 

\paragraph{Instruction Tuning for Multimodal Retrieval.}



Instruction-tuning is a popular strategy to enhance the multi-task capacity for both large language models~\cite{ouyang2022training,wei2021finetuned,chung2024scaling} and embedding models~\cite{su2022one,asai2022task,zhang2023retrieve,bge,bge-m3}. While there have been a few instruction datasets proposed for multimodal retrieval~\cite{cirr-liu2021image,univl-liu2022universal,chang2022webqa,mbeir-wei2023uniir,vista}, they are limited in scale and diversity due to their reliance on human annotation. Recently, a notable progress was made by MagicLens~\cite{magiclens}, where a large-scale open-ended search instruction dataset is created from the co-existed images within webpages. However, given the shortage of multi-image webs, MagicLens is limited by its scalability and data-quality. Moreover, this dataset is still held private and inaccessible to public users. As a result, the creation and release of high-quality instruction-tuning datasets have become imperative for advancing multimodal retrieval research.

\section{Methodology} 

\subsection{MegaPairs Construction}

Training on large-scale open-world data significantly enhances the generalization capabilities of foundation models.  
For instance, CLIP~\cite{clip-radford2021learning} has achieved remarkable advancements in cross-modal retrieval and various downstream tasks due to its extensive training on text-image pairs. However, the multimodal instruction tuning data, despite its importance to multimodal retrieval \cite{magiclens}, is scarce in natural world and expensive to annotate by human effort. In this paper, we propose to construct large-scale multimodal instruction-tuning datasets through data synthesis. Formally, each data instance contains the following triplet: a pair of images \((\mathcal{I}_q, \mathcal{I}_t)\), together with a textual instructions \(\mathcal{T}_{q \rightarrow t}\) specifying the transition relationship from query image $\mathcal{I}_q$ to target image $\mathcal{I}_t$. 


We identify two primary technical challenges in acquiring such triplets: (1) \textit{sampling relevant and diversified image pairs at scale}, (2) \textit{precise annotation of instruction for the sampled image pair}. To address these challenges, we propose leveraging the common open-domain image corpora. Intuitively, a large-scale corpus contains abundant correlated images of diverse semantic relationships, which can be mined and annotated for our instruction-tuning data. Our data synthesis pipeline is demonstrated as \Cref{fig:pipeline}, which involves two main components: the mining of image pairs and the generation of open-ended instructions. 


\paragraph{Mining Correlated Image Pairs.} 

As illustrated in \Cref{fig:pipeline}(a), we propose sampling correlated image pairs from a large-scale image corpus. For each query  image \((\mathcal{I}_q, \mathcal{C}_q)\), we utilize multiple similarity models to search for a diverse set of correlated target images of heterogeneous correlations \(\{\mathcal{I}_{t_1}, \mathcal{I}_{t_2}, \ldots, \mathcal{I}_{t_n}\}\). In our work, the following types of correlations are used: (1) visual-semantic correlation, which measures the semantic correlation of two images regardless of visual similarity, e.g., two different views of the same cars; (2) visual-pattern correlation, which captures the visual similarity of two images regardless of semantic correlation, e.g., different cars in similar backgrounds; (3) caption correlation, which measures the textual similarity between two images' captions. 


Recognizing the importance of \textbf{hard negatives} in training retrieval models \cite{xiong2020approximate,hofstatter2021efficiently,zhang2022uni}, for each pair \((\mathcal{I}_q, \mathcal{I}_{t_i})\), we include additional images \(\{\mathcal{I}_{t_j} \mid t_j \neq t_i\}\) from the retrieved set as hard negative samples. This approach is simple but empirically effective. We validate the scalability and quality of our data pairs in \Cref{subsec:exp-abl}, with additional examples visualized in \Cref{appendix-vis-example}. 

\paragraph{Generating Open-Ended Instructions.}

As shown in \Cref{fig:pipeline}(b), we utilize open-source multimodal large language models (MLLM) and large language models (LLM) for the automated annotation of mined image pairs \(\mathbf{\mathcal{P}} = \{(\mathcal{I}_q, \mathcal{I}_{t_i})\}\). Initially, each image pair \((\mathcal{I}_q, \mathcal{I}_{t_i})\) is processed by the MLLM to generate a detailed description \(\mathcal{D}_i\) of the common concepts and differences between the query image \(\mathcal{I}_q\) and the target image \(\mathcal{I}_{t_i}\). This description \(\mathcal{D}_i\) is then refined by the LLM to produce textual instructions \(\mathcal{T}_{q \rightarrow t_i}\). We prompt the LLM to generate multiple \(\mathcal{T}_{q \rightarrow t_i}\) for each pair, enhancing the diversity of the textual instructions. Ultimately, we construct a multimodal triplet \((\mathcal{I}_q, \mathcal{T}_{q \rightarrow t_i}, \mathcal{I}_{t_i})\), where \((\mathcal{I}_q, \mathcal{T}_{q \rightarrow t_i})\) can be used to retrieve \(\mathcal{I}_{t_i}\). This two-step annotation method ensures both accuracy and diversity in the automated annotation process while leveraging open-source models. The detailed prompts can be found in \Cref{appendix-data-construct}.

\paragraph{Implementations.}

A dataset of 26,235,105 image pairs is created based on the above data synthesis pipeline. We utilize a subset from the Recap-DataComp-1B~\cite{recap-datacomp} as our image corpus, containing 20 million captioned images. For similarity models, we employ EVA-CLIP's image encoder for visual-semantic correlation~\cite{evaclip-sun2023eva}, DINOv2~\cite{dinov2} for visual-pattern correlation, and EVA-CLIP's text encoder for caption similarity. We filter the image pairs whose similarity score is within (0.8, 0.96), thus eliminating weak associations and near duplications. We further leverage InternVL2-26B~\cite{internvl2} and LLaMA3-8B~\cite{llama3} to generate the open-ended instructions. For each image pair, we create at least three different textual instructions and introduce five hard negatives. 

\subsection{MMRet Model}
We propose MMRet, a series of models designed for universal multimodal retrieval based on pre-trained vision-language models (VLMs). Our MMRet integrates two distinct VLM architectures to achieve a universal multimodal embedding.

\paragraph{CLIP-based MMRet. }

The original CLIP~\cite{clip-radford2021learning} model employs a dual encoder architecture that independently encodes image and text data. We denote the image encoder as \({\Phi}_{{I}}\) and the text encoder as \({\Phi}_{{T}}\). Given an image \(I\) or text \(T\), their embeddings are computed as follows:
\begin{equation}
\begin{aligned}
\mathbf{e}_i &= {\Phi}_{{I}} (I) \\
\mathbf{e}_t &= {\Phi}_{{T}} (T)
\end{aligned}
\label{eq:clip-unimodal}
\end{equation}
To produce the multimodal embedding for a composed image-text sample \((I, T)\), we employ the score-fusion strategy as used by UniIR \cite{mbeir-wei2023uniir}, which directly uses an element-wise addition of the outputs from the dual encoders:
\begin{equation}
\mathbf{e}_{it} = {\Phi}_{{I}} (I) + {\Phi}_{{T}} (T)
\label{eq:clip-multimodal}
\end{equation}
In our CLIP-based MMRet, we trained both base and large models.


\paragraph{MLLM-based MMRet. }

The multimodal large language models (MLLMs) incorporate a visual encoder, typically based on a vision transformer~\cite{vit-DBLP:conf/iclr/DosovitskiyB0WZ21}, into a large language model (LLM). This integration allows image tokens to be directly processed by the LLM. Consequently, MLLMs can effectively handle diverse multimodal inputs by converting any type of input into a sequence of tokens. For instance, composed image-text data is transformed into interleaved sequences of image and text tokens, enabling the model to process them seamlessly.

Our MMRet model builds upon the LLaVA-1.6~\cite{llava-next}. In both training and inference stages, MMRet uses task-specific instructions for query inputs to improve generalization, aligning with standard practices in LLM-based embedding models~\cite{e5-mistral,llama2vec}. A typical multimodal query input is structured as follows:
\begin{equation}
\langle\text{instruct}\rangle ~~ \text{\{task\_inst\}} ~~ \langle\text{query}\rangle ~~ \{q_t\} ~~ \{q_i\} ~~\texttt{[EOS]}
\label{eq:mllm-input}
\end{equation}
where \(\text{\{task\_inst\}}\) represents the task-specific instruction, \(\{q_t\}\) denotes the input query text, and \(\{q_i\}\) is the input query image. The normalized last hidden state of the \texttt{[EOS]} token in the MLLM is used as the embedding of any given input sequence.

\subsection{Multimodal Contrastive Learning}

We employ multimodal contrastive learning to transform the original CLIP and MLLM into our MMRet model, enabling various multimodal retrieval tasks. We use the standard InfoNCE loss~\cite{infonce} as our training objective:
\begin{equation}
    \mathcal{L} = -\frac{1}{|\mathcal{Q}|} \sum_{q_i \in \mathcal{Q}} \log \frac{\exp(\mathbf{e}_{q_i} \cdot \mathbf{e}_{c_i^{+}} / \tau)}{\sum_{c_j \in \mathcal{C}} \exp(\mathbf{e}_{q_i} \cdot \mathbf{e}_{c_j} / \tau)}
    \label{eq:loss_function}
\end{equation}
where the set \( \mathcal{Q} \) includes all query samples \( q_i \) in a batch. The vectors \( \mathbf{e}_{q_i} \) and \( \mathbf{e}_{c_i^{+}} \) are the embeddings of the query \( q_i \) and its positive candidate \( c_i^{+} \), respectively. The set \(\mathcal{C}\) contains all in-batch candidates. Notably, both \( q \) and \( c \) can be images, text, or composed image-text data. The parameter \(\tau\) modulates the penalties on negative samples and is set to 0.02 unless otherwise specified in this paper.

\begin{table*}[h]
\begin{adjustbox}{width=\textwidth}
\begin{tabular}{l|cc|ccccc}
\toprule
\multirow{2}{*}{\centering \textbf{Methods}} & \multirow{2}{*}{\centering \textbf{Backbone}} & \multirow{2}{*}{\centering \textbf{\# Params}} & \multicolumn{1}{c}{\textbf{CIRCO}} & \multicolumn{2}{c}{\textbf{CIRR}} & \multicolumn{1}{c}{\textbf{FashionIQ}} & \multicolumn{1}{c}{\textbf{GeneCIS}} \\
\cmidrule(lr){4-4} \cmidrule(lr){5-6} \cmidrule(lr){7-7} \cmidrule(lr){8-8}
              &                   &           & mAP@5  & R@1 & R$_s$@1 & R@10 & R$_s$@1 \\\midrule
SEARLE~\cite{circo}        & CLIP-B            & 165M      & 9.4   & 24.0  & 54.9    & 22.9      & -       \\
CIReVL~\cite{CIReVL2024}  & CLIP-B            & 12.3B$^{\dag}$      & 14.9   & 23.9  & 60.2    & 28.3      & 15.9       \\
LDRE~\cite{ldre2024}          & CLIP-B            & 7.9B$^{\dag}$         & 18.0  & 25.7  & 60.5    & 24.8      & -       \\
MagicLens-B~\cite{magiclens}     & CLIP-B            & 166M      & 23.1  & 27.0  & 66.7    & 26.3      & 15.0    \\
\textcolor{gray}{MagicLens-B}$^{\ddagger}$~\cite{magiclens} & \textcolor{gray}{CoCa-B} & \textcolor{gray}{267M} & \textcolor{gray}{\underline{30.8}} & \textcolor{gray}{\underline{31.6}} & \textcolor{gray}{\underline{69.3}} & \textcolor{gray}{\textbf{35.2}} & \textcolor{gray}{\underline{17.4}}$^*$ \\
\midrule
\rowcolor{gray!20} 
\textbf{MMRet-Base} & CLIP-B     & 149M      & \textbf{34.3}  & \textbf{36.1}  &  \textbf{71.6} & \underline{31.9}      &    \textbf{18.0}     \\ 
\midrule
Pic2Word~\cite{saito2023pic2word}      & CLIP-L            & 429M      & 8.7   & 23.9  & - & 24.7      & 11.2    \\
PLI~\cite{PLI2023}           & CLIP-L            & 428M      & 10.4  & 25.5  & 55.6    & 35.4      & -       \\
SEARLE~\cite{circo}        & CLIP-L            & 442M      & 11.7  & 24.2  & 53.8 & 25.6      & 12.3    \\
CompoDiff~\cite{gu2023compodiff}     & CLIP-L            & 568M      & 12.6  & 18.2  & 57.4 & \underline{36.0}      & 14.9    \\
CIReVL~\cite{CIReVL2024}  & CLIP-L            & 12.5B$^{\dag}$      & 18.6   & 24.6  & 59.5    & 28.6      & 15.9       \\
LDRE~\cite{ldre2024}          & CLIP-L            & 8.2B$^{\dag}$         & 23.4  & 26.5  & 60.4    & 28.5      & -       \\
MagicLens-L~\cite{magiclens}     & CLIP-L            & 465M      & 29.6  & 30.1  & 68.1 & 30.7      & 16.3    \\
\textcolor{gray}{MagicLens-L}$^{\ddagger}$~\cite{magiclens} & \textcolor{gray}{CoCa-L} & \textcolor{gray}{613M} & \textcolor{gray}{\underline{34.1}}$^*$ & \textcolor{gray}{\underline{33.3}}$^*$ & \textcolor{gray}{\underline{70.9}}$^*$ & \textcolor{gray}{\textbf{38.0}} & \textcolor{gray}{\underline{16.7}} \\
\midrule
\rowcolor{gray!20} 
\textbf{MMRet-Large} & CLIP-L     & 428M      & \textbf{39.2}  & \textbf{38.0}  & \textbf{73.2}  & 34.6      &   \textbf{18.1}      \\ 
\midrule
LDRE~\cite{ldre2024}          & CLIP-G            & 10.3B$^{\dag}$         & 31.1  & 36.2  & 68.8    & 32.5      & -       \\
CIReVL~\cite{CIReVL2024}  & CLIP-G           & 14.6B$^{\dag}$      & 26.8  & 34.7  & 68.0    & 32.2      & \underline{17.4}$^*$      \\
IP-CIR~\cite{ip-cir2024}        & CLIP-G            & 43.8B$^{\dag}$         & \underline{32.8}  & \underline{39.3}  & \underline{70.0}    & \textbf{45.7}$^*$      & -       \\
E5-V~\cite{e5-v2024}          & LLaVA-1.6         & 8.35B     & 19.1  & 33.9  & -    & 31.8      & -       \\
MM-Emded~\cite{nv-mm-embed2024}      & LLaVA-1.6         & 7.57B     & 32.3  & -     & -    & -         & -       \\
\midrule
\rowcolor{gray!20} 
\textbf{MMRet-MLLM} & LLaVA-1.6   & 7.57B     & \textbf{42.2}  & \textbf{46.7}  & \textbf{75.4}  & \underline{35.6}      & \textbf{21.1}   \\ 
\bottomrule
\end{tabular}
\end{adjustbox}
\vspace{-0.2cm}
\caption{Zero-shot retrieval performance on various CIR benchmarks. $^*$ denotes the previous best performance for each benchmark prior to MMRet. $^\dag$ indicates methods with multiple components (e.g., GPT-3.5, Qwen1.5-32B); we report \# parameters of components with known sizes. The \textcolor{gray}{CoCa-based MagicLens}$^{\ddagger}$ models are proprietary. Results in \textbf{bold} and \underline{underline} denote the best and second-best performances for each model scale, respectively. Our MMRet model achieves state-of-the-art results across different model sizes and benchmarks, surpassing the previous SOTA by 8.1\% on the main benchmark CIRCO, significantly advancing zero-shot CIR methods.}
\vspace{-0.4cm}
\label{tab:cir-main}
\end{table*}

\section{Experiments}
\label{sec:exp}


In this section, we first evaluate the effectiveness of MegaPairs on zero-shot composed image retrieval (CIR) tasks in~\Cref{subsec:exp-zs-cir}. Next, we explore the impact of MegaPairs on broader multimodal retrieval tasks in~\Cref{subsec:exp-mmeb}. Finally, we conduct detailed analysis on our MegaPairs in~\Cref{subsec:exp-abl}.


\subsection{Zero-shot Performance on CIR tasks}
\label{subsec:exp-zs-cir}

\subsubsection{Implementation Details}
\label{subsubsec:zs-cir-imple}



We utilize our MegaPairs dataset to perform multimodal contrastive training for our MMRet models. For the CLIP-based MMRet, we initialize the model using both the base\footnote{https://huggingface.co/openai/clip-vit-base-patch16} and large\footnote{https://huggingface.co/openai/clip-vit-large-patch14} versions of CLIP, referred to as MMRet-Base and MMRet-Large, respectively. For the MLLM-based MMRet, we leverage the LLaVA-1.6 Mistral 7B architecture\footnote{https://huggingface.co/llava-hf/llava-v1.6-mistral-7b-hf} and initialize the model parameters accordingly, which we denote as MMRet-MLLM. The training details of MMRet on MegaPairs can be found in~\Cref{appendix-train-detail}.

\subsubsection{Benchmarks}
We evaluate our MMRet in a zero-shot setting across four different composed image retrieval benchmarks: CIRCO~\cite{circo}, CIRR~\cite{cirr-liu2021image}, FashionIQ~\cite{fashioniq-wu2021fashion}, and GeneCIS~\cite{genecis2023}. 
Following previous practice~\cite{magiclens}, CIRCO is considered our main benchmark due to its extensive candidate pool and high-quality annotations. Detailed information and metrics for each benchmark can be found in \Cref{appendix-eval-detail}.

\begin{table*}[h]
\centering
\begin{adjustbox}{width=\textwidth}
\begin{tabular}{l|cccc|c}
\toprule
\multirow{2}{*}{\textbf{Models}}  & \multicolumn{4}{c|}{\textbf{Per Meta-Task Score}}       & \multirow{2}{*}{\textbf{Overall}} \\
\cmidrule(lr){2-5}
     & Classification & VQA  & Retrieval & Grounding &                          \\
\midrule
\rowcolor{gray!10} 
\texttt{number of datasets}             & \texttt{10}             & \texttt{10}   & \texttt{12}        & \texttt{4}         & \texttt{36}                       \\
\midrule
BLIP2~\cite{blip2-li2023blip}                                                       & 27.0           & 4.2  & 33.9      & 47.0      & 25.2                     \\
SigLIP~\cite{siglip2023}                                                         & 40.3           & 8.4  & 31.6      & 59.5      & 34.8                     \\
CLIP~\cite{clip-radford2021learning}                                                        & 42.8           & 9.1  & 53.0      & 51.8      & 37.8                     \\
OpenCLIP~\cite{openclip2023}                                                & \textbf{47.8}           & 10.9 & 52.3      & 53.3      & 39.7                     \\
UniIR~\cite{mbeir-wei2023uniir}                                    & 42.1           & \underline{15.0} & \textbf{60.1}$^\dag$      & \textbf{62.2}      & \underline{42.8}                     \\
MagicLens~\cite{magiclens}                                                 & 38.8           & 8.3  & 35.4      & 26.0      & 27.8                     \\
E5-V (LLaVA-1.6)~\cite{e5-v2024}                                                           & 21.8           & 4.9  & 11.5      & 19.0      & 13.3                     \\
\midrule
\rowcolor{gray!20} 
\textbf{MMRet-MLLM (LLaVA-1.6)}                                                    & \underline{47.2}           & \textbf{18.4} & \underline{56.5}      & \textbf{62.2}      & \textbf{44.0}                     \\
\bottomrule
\end{tabular}
\end{adjustbox}
\vspace{-0.1cm}
\caption{Zero-shot performance on the Massive Multimodal Embedding Benchmark (MMEB). $^\dag$UniIR was trained on M-BEIR~\cite{mbeir-wei2023uniir}, which includes 10 of the 12 datasets in the MMEB retrieval tasks, it does not strictly adhere to a zero-shot setting. In contrast, our MMRet-MLLM, trained exclusively on the MegaPairs dataset, achieves state-of-the-art zero-shot performance in overall scores and multiple meta-tasks on MMEB.}
\vspace{-0.15cm}
\label{tab:zs-mmeb}
\end{table*}

\subsubsection{Evaluation Results}

The main evaluation results of MMRet across four benchmarks are shown in~\Cref{tab:cir-main}, with full results for each benchmark provided in~\Cref{appendix-full-CIR-results}. We have identified three key observations:

\textbf{(1) Our MMRet-MLLM model achieves leading performance across three of the four benchmarks.} Specifically, on our main benchmark, CIRCO, MMRet-MLLM surpasses the current SOTA CoCa-based MagicLens-L by achieving 42.2\% mAP@5 compared to 34.1\% (an increase of 8.1\%). On CIRR test set, it exceeds the current SOTA by 7.4\% and 4.5\% in R@1 and R$_s$@1, respectively. Additionally, on GeneCIS, it leads the current SOTA by 3.7\% in R$_s$@1.

\textbf{(2) MMRet exhibits superior performance across all model scales.} For instance, MMRet-Base and MMRet-Large outperform comparable models by 4.5\% and 4.7\% in R@1 on the CIRR test set, respectively. Additionally, they surpass similar models by 3.5\% and 5.1\% in mAP@5 on the CIRCO benchmark. In the fashion-domain benchmark FashionIQ, while not achieving the highest scores, our CLIP-based MMRet shows competitive performance against other CLIP-based models.

\textbf{(3) The MMRet-Base model surpasses most larger models, underscoring the exceptional quality of our MegaPairs dataset.} Despite being our smallest model, MMRet-Base outperforms many larger models such as the MagicLens-L. For instance, it achieving the best result on CIRCO with a mAP@5 of 34.3\%, excluding our own MMRet-Large and MMRet-MLLM models. It even exceeds the performance of models with dozens of times more parameters (e.g., MM-Embed), emphasizing the effectiveness of our MegaPairs dataset.

\subsection{Performance on MMEB}
\label{subsec:exp-mmeb}

To further validate the generalization ability of MegaPairs for broader multimodal embedding tasks, we evaluate MMRet on the Massive Multimodal Embedding Benchmark (MMEB)~\cite{vlm2vec2024}. MMEB is a comprehensive benchmark that includes 36 datasets across four meta-task categories: \textit{Classification}, \textit{Visual Question Answering}, \textit{Retrieval}, and \textit{Visual Grounding}. It is designed to evaluate the quality of multimodal embeddings and assesses models across diverse combinations of text and image modalities. We present the performance of MMRet in both zero-shot and supervised fine-tuning scenarios. Following previous works~\cite{e5-v2024,vlm2vec2024}, we conduct experiments using our MMRet-MLLM.

\begin{table*}[h]
\centering
\begin{adjustbox}{width=\textwidth}
\begin{tabular}{l|cccc|ccc}
\toprule
\multirow{2}{*}{\textbf{Models}}  & \multicolumn{4}{c|}{\textbf{Per Meta-Task Score}}       &\multicolumn{3}{c}{\textbf{Average Scor}e} \\
\cmidrule(lr){2-5}
\cmidrule(lr){6-8}
 & Classification & VQA  & Retrieval & Grounding &  IND  & OOD & Overall                     \\
\midrule
\rowcolor{gray!10} 
\texttt{number of datasets}      & \texttt{10}             & \texttt{10}   & \texttt{12}        & \texttt{4}     & \texttt{20}   & \texttt{16}       & \texttt{36}                       \\
\midrule
CLIP                     & 55.2           & 19.7 & 53.2      & 62.2      & 47.6     & 42.8     & 45.4        \\
OpenCLIP                 & \textbf{56.0}           & 21.9 & 55.4      & 64.1      & 50.5     & 43.1     & 47.2        \\
VLM2Vec (LLaVA-1.6)        & 54.7           & 50.3 & 56.2      & 64.0      & 61.0     & 47.5     & 55.0        \\
VLM2Vec (Phi-3.5-V)        & 54.8           & \underline{54.9} & \underline{62.3}      & \underline{79.5}      & \underline{66.5}     & \underline{52.0}     & \underline{60.1}        \\
\midrule
\rowcolor{gray!20} 
\textbf{MMRet-MLLM}               & \textbf{56.0}           & \textbf{57.4} & \textbf{69.9}      & \textbf{83.6}      & \textbf{68.0}     & \textbf{59.1}     & \textbf{64.1}  \\
\bottomrule
\end{tabular}
\end{adjustbox}
\vspace{-0.1cm}
\caption{Supervised fine-tuning results on the MMEB benchmark. The backbone of our MMRet-MLLM is LLaVA-1.6~\cite{llava-next}. We compare our results with the following baselines: CLIP~\cite{clip-radford2021learning}, OpenCLIP~\cite{openclip2023}, and two versions of VLM2Vec~\cite{vlm2vec2024} that employ the LLaVA-1.6~\cite{llava-next} and Phi-3.5-V~\cite{phi3} backbones. All baseline results are sourced from \cite{vlm2vec2024}. IND: in-distribution dataset; OOD: out-of-distribution dataset.}
\vspace{-0.1cm}
\label{tab:finetune-mmeb}
\end{table*}

\subsubsection{Zero-shot Performance}
\label{subsubsec:zs-mmeb}
\paragraph{Implementation Details. }

In the zero-shot evaluation on MMEB, we directly utilize our MMRet-MLLM from \Cref{subsec:exp-zs-cir}, maintaining implementation details consistent with \Cref{subsubsec:zs-cir-imple}.

\paragraph{Metrics.} We evaluate Precision@1 for all tasks, which measures the ratio of positive candidates ranked in the top position for all queries. We report the average scores for the four meta tasks as well as the overall average. Following the MMEB setting, we incorporate the predefined task-specific instructions into queries for all tasks during evaluation.

\paragraph{Results.} The zero-shot performance of our MMRet-MLLM on MMEB is presented in~\Cref{tab:zs-mmeb}. MMRet-MLLM achieved state-of-the-art zero-shot performance across various embedding meta-task, recording the highest overall average performance. Compared to the recent E5-V~\cite{e5-v2024}, which uses a similar LLaVA-1.6~\cite{llava-next} backbone for universal multimodal embedding, MMRet-MLLM trained on our MegaPairs dataset demonstrated superior performance. Notably, the second-best model, UniIR, was trained on M-BEIR~\cite{mbeir-wei2023uniir}, which encompasses datasets from 10 of the 12 retrieval meta-tasks in MMEB, and thus is not considered zero-shot for this meta-task. Consequently, our MLLM-Ret significantly outperforms the remaining methods in the retrieval meta-task and demonstrates strong generalization capabilities across all tasks.



\subsubsection{Supervised Fine-tuning Performance}
\paragraph{Implementation Details.}
We further fine-tune our MMRet-MLLM on MMEB to investigate the impact of MegaPairs on downstream task performance. The MMEB dataset includes 20 in-distribution (IND) datasets for training and 16 out-of-distribution (OOD) datasets for evaluation. We utilize the training sets from the 20 IND datasets, comprising approximately 662K data points. The learning rate is set to \(5 \times 10^{-6}\), and we employ LoRA with a rank of 32. The batch size is set to 192, and we train for one epoch. Following the VLM2Vec configuration~\cite{vlm2vec2024}, we incorporate task-specific instructions into the queries during training. 

\paragraph{Metrics.}
We employ the same metrics as outlined in~\Cref{subsubsec:zs-mmeb}. Additionally, we report the average scores for both the IND and OOD datasets.

\paragraph{Results. }

Table~\ref{tab:finetune-mmeb} compares the supervised fine-tuning performance of our MMRet model with various baselines on the MMEB dataset. Our MMRet-MLLM achieves state-of-the-art performance, with an overall average Precision@1 of 64.1\%. Compared to VLM2Vec (LLaVA-1.6)~\cite{vlm2vec2024}, which directly fine-tunes LLaVA-1.6 on MMEB, MMRet-MLLM enhances downstream task performance by 9.1\% through multimodal contrastive training on our MegaPairs. Notably, our model shows improvements of 11.6\% and 7.1\% on out-of-distribution (OOD) datasets compared to the two versions of VLM2Vec, highlighting the superior generalization capability of our MegaPairs for broader downstream multimodal embedding tasks.

\subsection{Detailed Investigation on MegaPairs}
\label{subsec:exp-abl}

We first assess the quality and scalability of our MegaPairs dataset in~\Cref{subsubsec:data-scale}. Next, we evaluate the effectiveness of the hard negative samples provided by MegaPairs in~\Cref{subsubsec:data-hn}. Finally, we explore the strategies used for mining image pairs from open-domain image corpora in~\Cref{subsubsec:data-search}. Unless otherwise specified, all subsequent experiments are conducted using our MMRet-base model.

\subsubsection{Data Scalability and Quality}
\label{subsubsec:data-scale}
\begin{figure}[tb]
    \centering
    \includegraphics[width=0.475\textwidth]{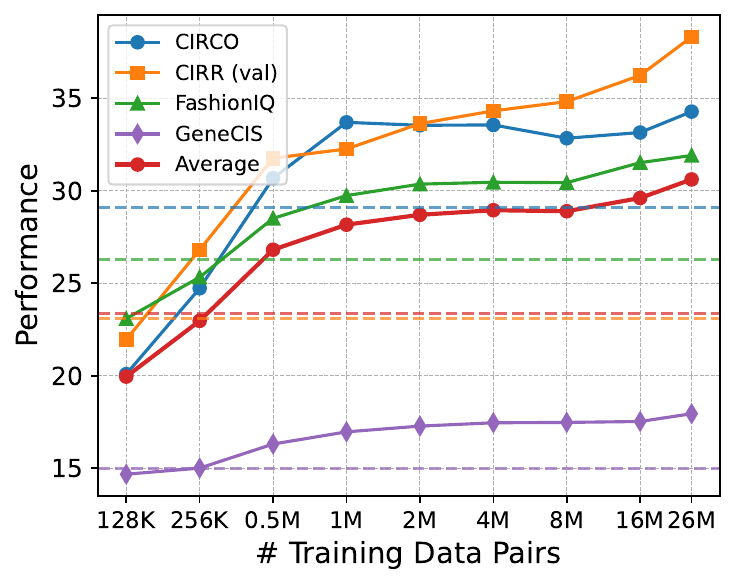}
    \vspace{-0.2cm}
    \caption{Performance scaling of MMRet-base on the MegaPairs as data size increases. The dashed lines indicate the performance of MagicLens-B (CLIP) trained on their dataset of 36.7M data pairs.}
    \label{fig:abl-scaling}
\end{figure} 
We first evaluated the performance trend of MMRet by training it on different sizes of subsets from the MegaPairs dataset to verify its scalability. Subsequently, we compared it with existing datasets to highlight the high-quality features of MegaPairs.

\noindent\textbf{Performance Scaling.} As shown in~\Cref{fig:abl-scaling}, the performance of MMRet-base across various benchmarks consistently improves with the increasing size of training data. This upward trend highlights the effectiveness and scalability of MegaPairs.

\noindent\textbf{Dataset Quality Comparison with Exsiting Datasets.} The dashed lines in~\Cref{fig:abl-scaling} represent the performance of the MagicLens-B (CLIP) model, trained on their 36.7M dataset~\cite{magiclens}. Remarkably, with only 0.5M samples from our MegaPairs dataset, constituting less than 2\% of MagicLens, MMRet significantly surpasses MagicLens across all benchmarks using the same CLIP-base backbone. This result underscores the superior quality and efficiency of our MegaPairs dataset.

\subsubsection{The Impact of Hard Negatives}
\label{subsubsec:data-hn}

In MegaPairs, images from the retrieved set that are not the target are marked as hard negatives, providing a diverse and ample set of hard negative target images for each image pair. As shown in~\Cref{tab:abl-hn}, compared to not using negatives or only using the query image as a negative, training with our mined hard negatives significantly enhances model performance across all benchmarks.

\begin{table}[t]
\centering
\begin{adjustbox}{width=0.475\textwidth}
\begin{tabular}{cc|cccc}
\toprule
\multicolumn{2}{c|}{\textbf{Negatives}} & \textbf{CIRCO} & \textbf{CIRR}$^\dag$ & \textbf{FIQ} & \textbf{CIS} \\ 
\cmidrule(lr){1-2}
\cmidrule(lr){3-3}
\cmidrule(lr){4-4}
\cmidrule(lr){5-5}
\cmidrule(lr){6-6}
\textit{\textbf{Qry}} & \textbf{\textit{HN}} & mAP@5 & R@1 & R@10 & R$_s$@1 \\ 
\midrule
\crossmark  & \crossmark & 10.1  & 0.2 & 25.3 & 14.4 \\ 
\checkmark  & \crossmark & 29.7 & 32.1 & 27.6 & 16.6 \\ 
\checkmark  & \checkmark & \textbf{32.3} & \textbf{33.7} & \textbf{30.1} & \textbf{17.0} \\ 
\bottomrule
\end{tabular}
\end{adjustbox}
\vspace{-0.2cm}
\caption{Performance comparison of MMRet-base using different negative strategies at a 1M scale. \textit{\textbf{Qry}}: query image negative; \textit{\textbf{HN}}: our mined hard negatives. $^\dag$We report CIRR validation set performance due to their test server submission limits.}
\vspace{-0.2cm}
\label{tab:abl-hn}
\end{table}

\subsubsection{Data Pair Search Strategy}
\label{subsubsec:data-search}

We explored the impact of various search strategies in constructing heterogeneous triplets. For a fair comparison, we selected 1M data entries for each construction strategy and trained the model for 2000 steps.

~\Cref{tab:abl-pair} presents the results of various data pairing strategies across multiple benchmarks. Initially, when evaluating individual strategies, we observed that triplets based on text similarity achieved the highest zero-shot CIR performance. We hypothesize that text similarity captures more diverse relationships than image similarity. Furthermore, combining any two pairing strategies consistently outperformed using a single strategy. This enhancement is likely due to the increased diversity within the dataset, which is essential for training robust multimodal embedding models. Ultimately, employing all three strategies simultaneously provided the most robust performance across all datasets. As a result, this approach was chosen for constructing the MegaPairs.

\begin{table}[t]
\centering
\begin{adjustbox}{width=0.475\textwidth}
\begin{tabular}{ccc|cccc}
\toprule
\multicolumn{3}{c|}{\textbf{Strategy}} & \textbf{CIRCO} & \textbf{CIRR}$^\dag$ & \textbf{FIQ} & \textbf{CIS} \\ 
\cmidrule(lr){1-3}
\cmidrule(lr){4-4}
\cmidrule(lr){5-5}
\cmidrule(lr){6-6}
\cmidrule(lr){7-7}
\textit{\textbf{D}} & \textit{\textbf{I}} & \textit{\textbf{T}} & mAP@5 & R@1  & R@10 & R$_s$@1 \\ 
\midrule
\checkmark & \crossmark & \crossmark & 29.0 & 31.5 & 24.7 & 17.2 \\ 
\crossmark & \checkmark & \crossmark & 30.0 & 30.0 & 29.6 & 15.3 \\ 
\crossmark & \crossmark & \checkmark & 31.6 & 32.2 & 28.7 & \underline{17.3} \\ 
\midrule
\checkmark & \checkmark & \crossmark & 31.0 & 32.1 & 28.5 & 17.1\\ 
\checkmark & \crossmark & \checkmark & \textbf{32.4} & \underline{33.3} & 28.9 & \textbf{17.5} \\ 
\crossmark & \checkmark & \checkmark & 32.2 & \underline{33.3} & \underline{29.7} & 16.4 \\ 
\midrule
\checkmark & \checkmark & \checkmark & \underline{32.3} & \textbf{33.7} & \textbf{30.1} & 17.0 \\ 
\bottomrule
\end{tabular}
\end{adjustbox}
\caption{Performance comparison of MMRet-base using different data pairing strategies at 1M scale. \textit{\textbf{D}}: DINOv2 Encoder; \textit{\textbf{I}}: CLIP Image Encoder; \textit{\textbf{T}}: CLIP Text Encoder. FIQ and CIS represent the FashionIQ and GeneCIS benchmarks, respectively. $^\dag$ We report CIRR validation set performance due to test server submission limits.}
\label{tab:abl-pair}
\end{table}

\section{Conclusion}
In this paper, we introduce MegaPairs, a large-scale multimodal pairing dataset designed for training universal multimodal retrievers. MegaPairs comprises diverse image pairs from the open world, annotated with open-ended textual instructions that capture their visual and semantic relationships. Using MegaPairs, we trained our MMRet models, achieving state-of-the-art zero-shot performance in four composed image retrieval tasks and on the Massive Multimodal Embedding Benchmarks, which consists of 36 different datasets. Extensive experiments further demonstrate the generalization capability and high-quality features of MegaPairs.

\clearpage

\section*{Limitations}
In constructing MegaPairs, we discovered that using diverse retrievers can generate richer image pairs. Our study employed three distinct retrievers, which offered substantial diversity. However, there remains potential to explore additional pairing methods, such as leveraging more advanced text domain retrievers (e.g., BGE~\cite{bge}) or incorporating varied strategies like image-text retrieval.

\section*{Ethics Statement}
All images in our MegaPairs dataset are sourced from the Recap-Datacomp-1B dataset~\cite{recap-datacomp}, and have undergone rigorous screening by the Datacomp team to remove harmful content~\cite{datacomp}. Despite our best efforts, we acknowledge that these screenings may not be entirely comprehensive or without omissions. Additionally, we strongly discourage the use of MMRet models for encoding and retrieving sensitive content.

\bibliography{custom}

\appendix
\section*{Appendix}

\section{Detailed Prompt for Annotating Open-Ended Instructions}
\label{appendix-data-construct}
\begin{figure}[h]
    \centering
    \includegraphics[width=0.5\textwidth]{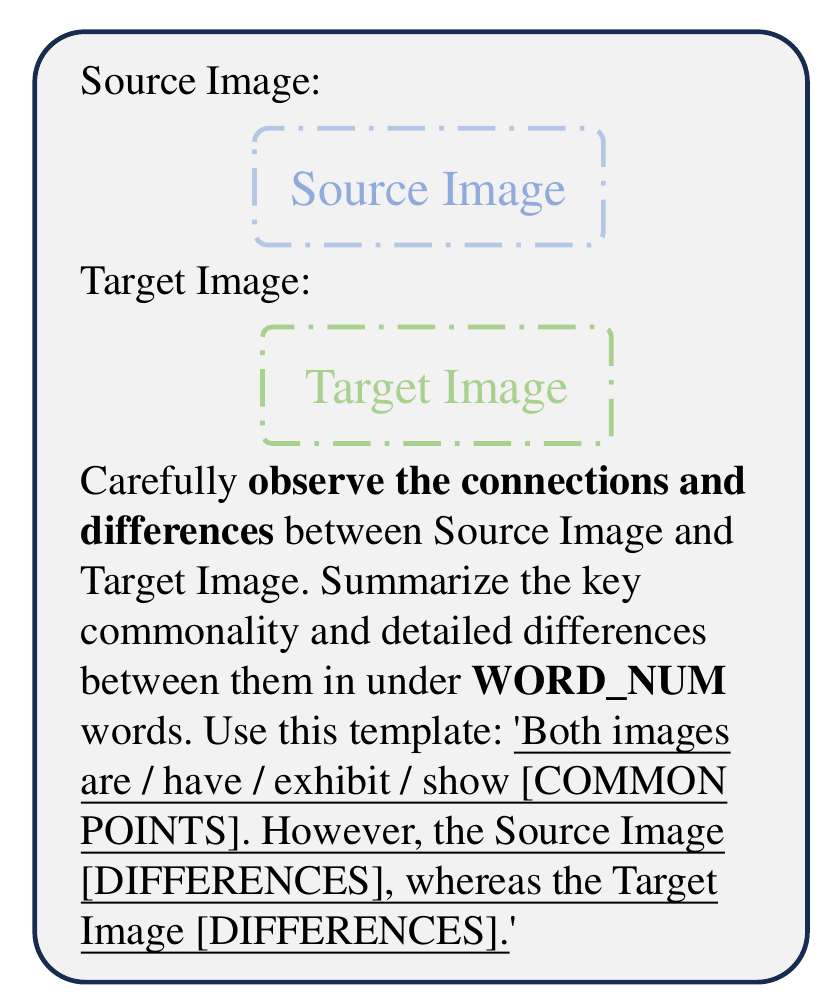}
    \vspace{-0.5cm}
    \caption{The specific prompts for MLLM. The value of WORD\_NUM ranges from 60 to 100 in our practical data generation to enhance the diversity of the generated description.}
    \vspace{-0.6cm}
    \label{fig:MLLM prompt examples}
\end{figure}

To annotate open-ended instructions, we begin by using the MLLM to generate a detailed description of the commonalities and differences between the query image and the target image, where the corresponding prompt is illustrated in Figure \ref{fig:MLLM prompt examples}. Subsequently, the description is refined by the LLM to produce textual instructions, with the associated prompt provided in Figure \ref{fig:LLM prompt examples}.

\begin{figure*}[tb]
    \centering
    \includegraphics[width=\textwidth]{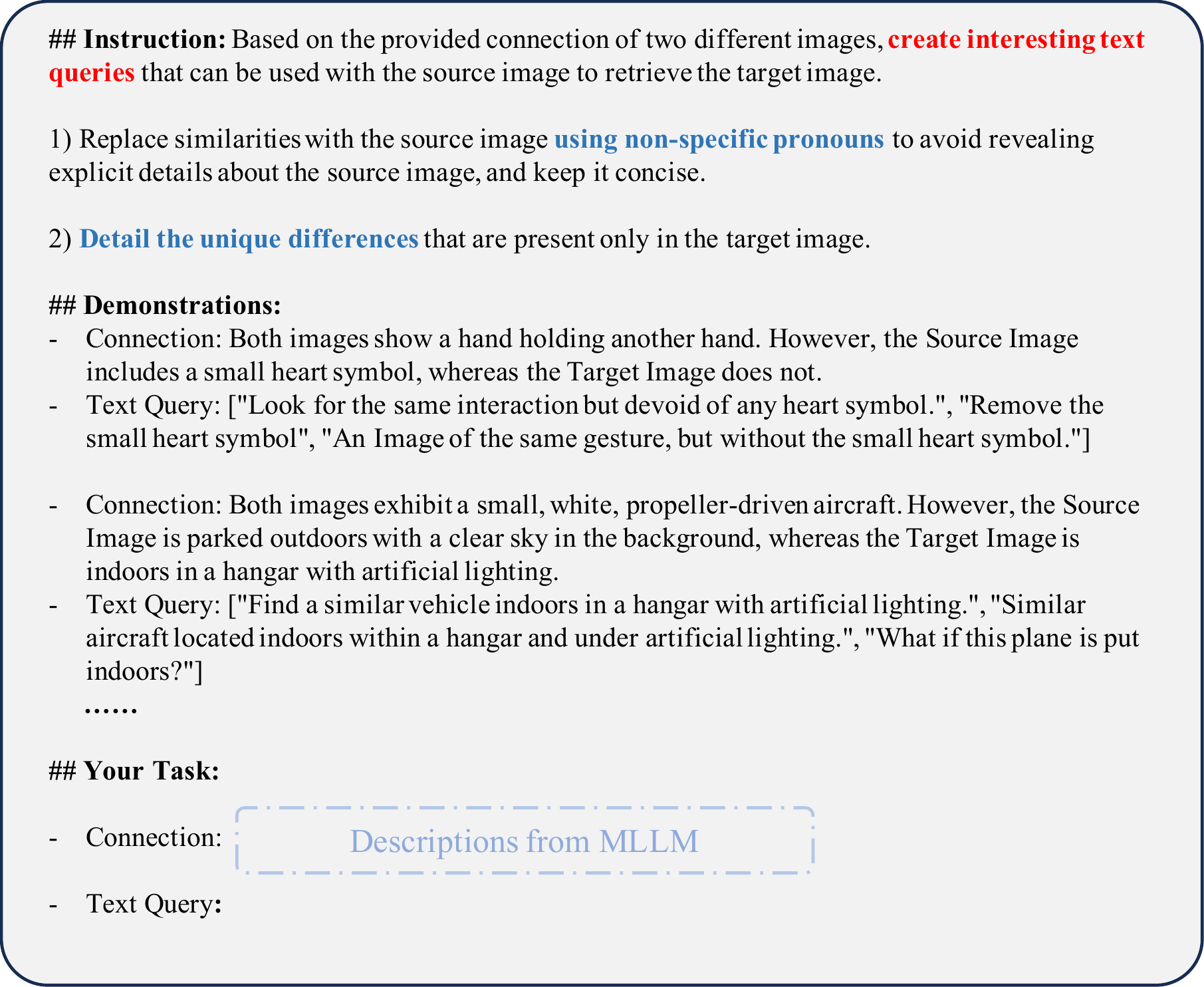}
    \vspace{-0.8cm}
    \caption{The specific prompts for LLM.  The figure showcases two demonstrations, while in our practical data generation process, five demonstrations are randomly selected from a pool of 50 and fed into the LLM. }
    \vspace{-0.6cm}
    \label{fig:LLM prompt  examples}
\end{figure*}

\section{Training Details of MMRet on MegaPairs}
\label{appendix-train-detail}
For the CLIP-based MMRet, the training process employs a batch size of 2048, with each query paired with one positive image and four hard negatives. All input images are resized to 224x224 to match the model's configuration. During training, all CLIP parameters remain unfrozen. MMRet-base is trained for 15,000 steps, while MMRet-large is trained for 25,000 steps on the MegaPairs dataset.

For the MLLM-based MMRet, we use a batch size of 144 during training, with each query associated with one positive image and three hard negatives. We apply LoRA~\cite{lora} to both the ViT encoder and the LLM backbone of LLaVA-1.6, setting the LoRA rank to 32. Although the original model supports variable resolution image processing, we use a fixed resolution of 512x512 for all images to manage the token sequence length. MMRet-MLLM is trained for 20,000 steps on the MegaPairs dataset.

For both CLIP-based and MLLM-based MMRet models, we set an initial learning rate of \(5 \times 10^{-6}\) and employ a linear decay strategy.

\section{Detailed Information and Evaluation Metrics of Zero-Shot CIR Benchmarks}
\label{appendix-eval-detail}
The detailed information and metrics of our evaluation in zero-shot composed image retrieval (CIR) tasks for each benchmark are as follows:

\noindent\textbf{CIRCO}~\cite{circo} is a challenging zero-shot CIR benchmark comprising 123,403 candidate natural images. We evaluete our MMRet models on its test set, which contains 800 composed image-text queries, each annotated with multiple ground-truth images. We use mean Average Precision (mAP) as the evaluation metric. Due to its extensive candidate pool and high-quality annotations, CIRCO serves as a robust and comprehensive benchmark for zero-shot CIR evaluation, and we consider it our main benchmark.

\noindent\textbf{CIRR}~\cite{cirr-liu2021image} is the first dataset for conducting the CIR task using natural images. We conduct zero-shot evaluations on its test set, which comprises 4,148 queries and a corpus of 2,315 images. Each query in CIRR is annotated with exactly one positive target image, but it suffers from some false negative issues. For each query, CIRR provides a subset retrieval setting that retrieves target images from a small corpus. We assess both standard and subset retrieval performance using recall metrics (R and R$_s$).

\noindent\textbf{FashionIQ}~\cite{fashioniq-wu2021fashion} is another CIR task focusing on fashion products. We conduct zero-shot evaluations on its validation set, which includes 6,016 queries and 15,536 images. FashionIQ comprises three sub-tasks: dress, shirt, and toptee. We evaluate each sub-task separately and report their average recall values.

\noindent\textbf{GeneCIS}~\cite{genecis2023}
is a benchmark for conditional image similarity measurement, comprising four sub-tasks about changing or focusing the attribute or object in the given image. 
In each sub-task, models need to retrieve the most similar images from a dedicated small subset for the given query image and the condition keyword.
We approach it as a CIR task by combining the query image and the text description of the sub-task derived from the condition keyword as a composed image-text query.
Each query's candidate subset averages $13.8$ images, and the mean R$_{s}$ across all four subsets is reported.

\section{Full results on CIR Benchmarks}
\label{appendix-full-CIR-results}
We report the full results on four CIR benchmarks ~\cite{circo, cirr-liu2021image, fashioniq-wu2021fashion, genecis2023} in Tables \ref{tab:CIRCO-full-results}, \ref{tab:CIRR-full-results}, \ref{tab:FashionIQ-full-results}, and \ref{tab:GeneCIS-full-results}, respectively. Our MMRet model achieves state-of-the-art performance across various model sizes on the CIRCO, CIRR, and GeneCIS benchmarks.

\section{Full Results on MMEB Benchmark}
\label{appendix-detail-results}
We list the full results on the MMEB benchmark~\cite{vlm2vec2024} in Table \ref{tab:mmeb_full_performance}. The MMEB benchmark consists of 36 datasets spanning four meta-task categories, including 20 in-distribution datasets and 16 out-of-distribution (OOD) datasets. The results on the OOD datasets are highlighted with a gray background in the table. Our MMRet model achieves state-of-the-art performance in both zero-shot and fine-tuning settings. Notably, MMRet surpasses the second-best performance on the OOD datasets, demonstrating its remarkable generalization capability.

\section{Visualized Examples of MegaPairs}
\label{appendix-vis-example}
\begin{figure*}[tb]
    \centering
    \includegraphics[width=\textwidth]{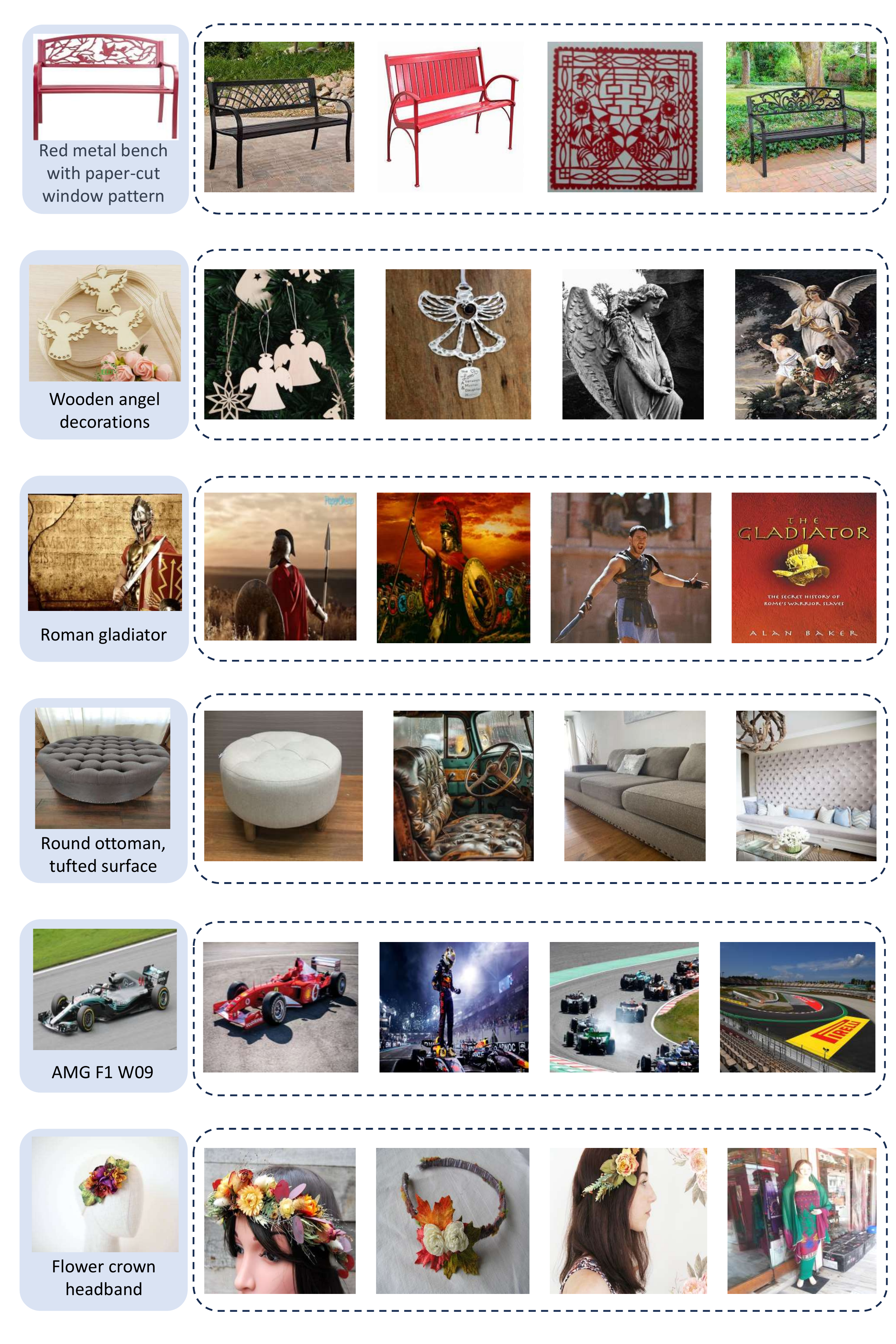}
    \vspace{-0.5cm}
    \caption{The visualized examples of MegaPairs. Each row represents a single example, with the query item highlighted in a blue rectangle and the target items enclosed within a dashed box.}
    \vspace{-0.6cm}
    \label{fig:example of datasets}
\end{figure*}

We present several examples of MegaPairs in Figure~\ref{fig:example of datasets}. Each row corresponds to a single example, where the query item, comprising an image and its corresponding alt-text caption, is associated with multiple target images. These target images include both visually similar ones and semantically related ones beyond visual features. 

For example, in the 4th row, the query image showing an ottoman with the alt-text caption \texttt{Round ottoman, tufted surface} is paired with target items that feature visually similar images (e.g., the 1st target image, which shows an ottoman, and the 3rd target image, which depicts a sofa with a similar style) as well as semantically related images that transcend visual features (e.g., the 2nd and 4th target images, depicting the interior of a car and a living room wall, respectively. These share few visual features with the query image but also exhibit a tufted surface). In the 5th row, the query image showing an F1 car with the alt-text caption \texttt{AMG F1 W09} is paired with target items featuring visually similar images (e.g., the 1st target image, which shows an F1 car in red, and the 3rd target image, which displays a race scene with multiple F1 cars) as well as semantically related images that transcend visual features (e.g., the 2nd target image, which shows an F1 driver, and the 4th target image, depicting an F1 circuit. These images bear no visual similarity to the query image but share the F1 concept).

\section{Qualitative Results of MMRet on Zero-shot CIR Tasks}
\begin{figure*}[tb]
    \centering
    \includegraphics[width=\textwidth]{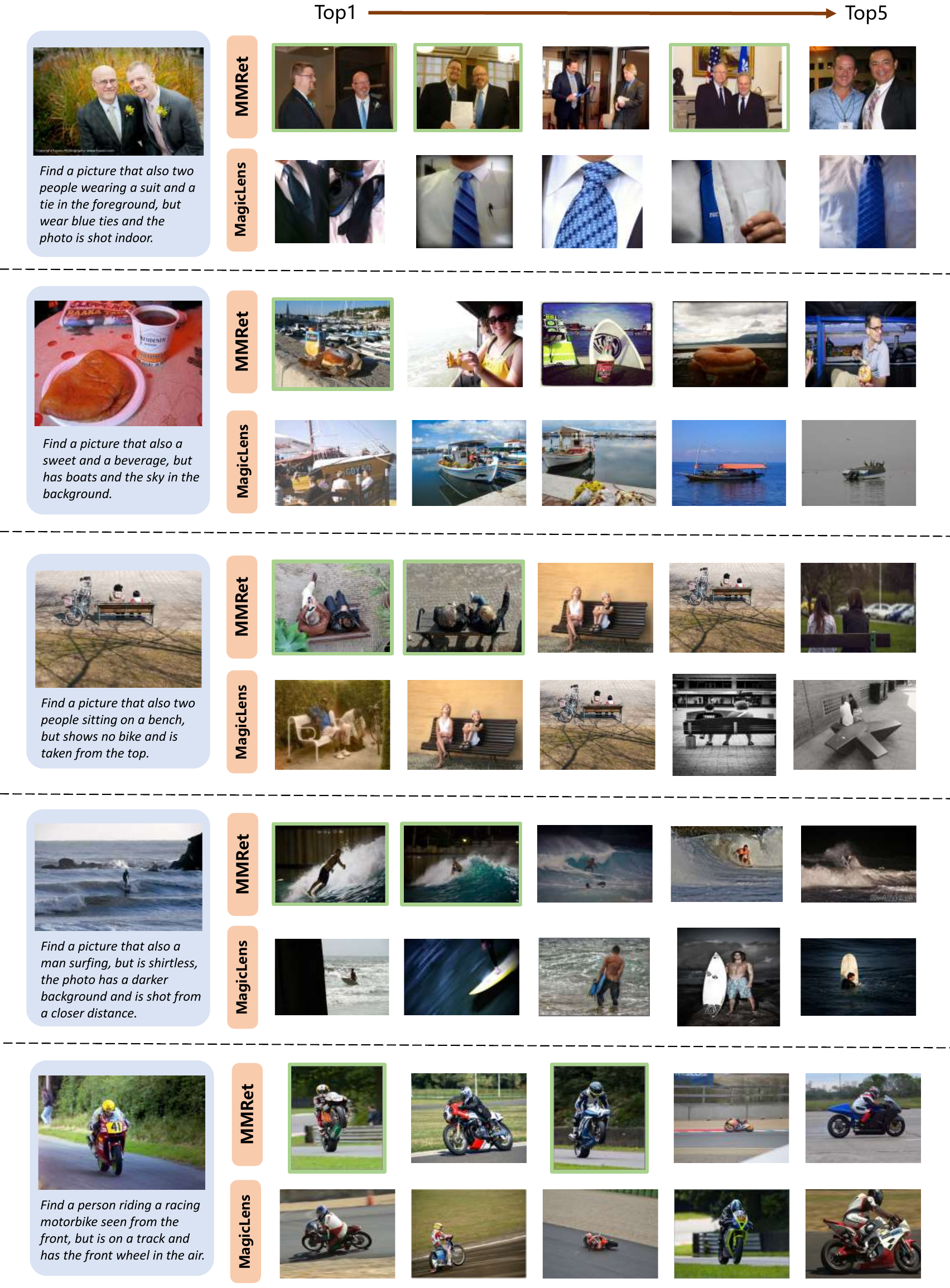}
    \vspace{-0.5cm}
    \caption{Top-5 retrieved images of MMRet and MagicLens on zero-shot CIR tasks, both using the CLIP-L backbone. Queries are shown with a blue background, and the most correct retrieved images are marked with green outlines.}
    \vspace{-0.6cm}
    \label{fig:ZS-CIR examples}
\end{figure*}

We present several top-5 retrieved images of our MMRet and the SOTA MagicLens~\cite{magiclens} on zero-shot CIR tasks, as shown in Figure \ref{fig:ZS-CIR examples}.  Since only the CLIP-based checkpoint is available for MagicLens, we select the CLIP-L backbone for both methods. 1) For the \texttt{blue ties} query, MMRet accurately interprets the query and identifies both the specific attire and indoor setting, retrieving multiple images that meet the specified requirements. In contrast, MagicLens focuses solely on the individual object, overlooking the broader semantic context. 2) For the \texttt{sweet, beverage, boats and sky} query, MMRet demonstrates a solid understanding of real-world entity concepts, successfully integrating both foreground and background elements to retrieve the most relevant image. 3) The success on the \texttt{bench top} query highlights MMRet's ability to comprehend specific pose and angle requirements. 4) The success on the \texttt{darker ground and closer distance} query illustrates MMRet's capacity to recognize lighting conditions and shooting distance. 5) The success on the \texttt{whell in the air} query indicates that MMRet can identify dynamic actions and contextual scene elements.

\begin{table*}[h]
\begin{adjustbox}{width=\textwidth}
\begin{tabular}{l|cc|cccc}
\toprule
\multirow{2}{*}{\centering \textbf{Methods}} & \multirow{2}{*}{\centering \textbf{Backbone}} & \multirow{2}{*}{\centering \textbf{\# Params}} & \multirow{2}{*}{mAP@5} & \multirow{2}{*}{mAP@10} & \multirow{2}{*}{mAP@25} & \multirow{2}{*}{mAP@50} \\
              &                   &           &   &  &  &  \\\midrule
PALAVRA~\cite{cohen2022my} & CLIP-B &176M &4.6 &5.3 &6.3 &6.8\\
PLI~\cite{PLI2023}           & BLIP-B            & 224M      & 7.1   & 8.0  & 9.2    & 9.7          \\
SEARLE~\cite{circo}        & CLIP-B            & 165M      & 9.4   & 9.9 &11.1 &11.8 \\
CIReVL~\cite{CIReVL2024}  & CLIP-B            & 12.3B$^{\dag}$      & 14.9 &15.4 &17.0 &17.8  \\
LDRE~\cite{ldre2024}          & CLIP-B            & 7.9B$^{\dag}$         & 18.0  & 18.3  & 20.2    & 21.1             \\
MagicLens-B~\cite{magiclens}     & CLIP-B            & 166M      & 23.1 &23.8 &25.8 &26.7       \\
\textcolor{gray}{MagicLens-B}$^{\ddagger}$~\cite{magiclens} & \textcolor{gray}{CoCa-B} & \textcolor{gray}{267M} & \textcolor{gray}{30.8} & \textcolor{gray}{32.0} & \textcolor{gray}{34.5} & \textcolor{gray}{35.6} \\
\midrule
\rowcolor{gray!20} 
\textbf{MMRet-Base} & CLIP-B     & 149M      & \textbf{34.3}  & \textbf{35.0}  &  \textbf{37.6} & \textbf{38.7}         \\ 
\midrule
Pic2Word~\cite{saito2023pic2word}      & CLIP-L            & 429M      & 8.7 &9.5 &10.6 &11.3       \\
PLI~\cite{PLI2023}           & CLIP-L            & 428M      & 10.4 &11.6 &13.0 &13.7    \\
SEARLE~\cite{circo}        & CLIP-L            & 442M      & 11.7 &12.7 &14.3 &15.1      \\
CIReVL~\cite{CIReVL2024}  & CLIP-L     & 12.5B$^{\dag}$    & 18.6 &19.0 &20.9 &21.8  \\
LinCIR~\cite{lincir2024}        & CLIP-L            & 442M      & 12.6 &13.6 &15.0 &15.9        \\
CompoDiff~\cite{gu2023compodiff}     & CLIP-L            & 568M      & 12.6 &13.4 &15.8 &16.4       \\
MagicLens-L~\cite{magiclens}     & CLIP-L            & 465M      & 29.6 & 30.8 & 33.4 & 34.4         \\
\textcolor{gray}{MagicLens-L}$^{\ddagger}$~\cite{magiclens} & \textcolor{gray}{CoCa-L} & \textcolor{gray}{613M} & \textcolor{gray}{34.1} & \textcolor{gray}{35.4} & \textcolor{gray}{38.1} & \textcolor{gray}{39.2}  \\
\midrule
\rowcolor{gray!20} 
\textbf{MMRet-Large} & CLIP-L     & 428M      & \textbf{39.2}  & \textbf{40.2}  & \textbf{42.9}  & \textbf{44.0}            \\ 
\midrule
Pic2Word~\cite{saito2023pic2word} & CLIP-H &987M &11.7 &12.3 &13.7 &14.4\\
SEARLE~\cite{circo} & CLIP-H &1.0B & 16.1 &16.9 &18.8 &19.7\\
LinCIR~\cite{lincir2024}        & CLIP-H     & 1.0B      & 17.6 &18.5 &20.5 &21.4\\
Pic2Word~\cite{saito2023pic2word} & CLIP-G &2.5B &5.5 &5.6 &6.7 &7.1\\
SEARLE~\cite{circo} & CLIP-G &2.6B & 13.2 &13.9 &15.3 &16.0\\
CompoDiff~\cite{gu2023compodiff} &CLIP-G &2.9B & 15.3 &17.7 &19.4 &- \\
CIReVL~\cite{CIReVL2024} &CLIP-G &14.6B$^{\dag}$ &26.8 &27.6 &30.0 &31.0 \\
LinCIR~\cite{lincir2024}        & CLIP-G            & 2.6B      & 19.7 &21.0 &23.1 &24.2\\
LDRE~\cite{ldre2024}          & CLIP-G            & 10.3B$^{\dag}$         & 31.1  & 32.2 &35.0 &36.0  \\
IP-CIR~\cite{ip-cir2024}        & CLIP-G            & 43.8B$^{\dag}$         & 32.8  &34.3 &36.9 &38.0    \\
E5-V~\cite{e5-v2024}          & LLaVA-1.6         & 8.35B     & 19.1  & -  &  -   & - \\
MM-Emded~\cite{nv-mm-embed2024}      & LLaVA-1.6         & 7.57B     & 32.3  &  -    & -    & -       \\
\midrule
\rowcolor{gray!20} 
\textbf{MMRet-MLLM} & LLaVA-1.6   & 7.57B     & \textbf{42.2}  & \textbf{43.4}  & \textbf{46.5}  & \textbf{47.6}  \\ 
\bottomrule
\end{tabular}
\end{adjustbox}
\caption{Full results on the CIRCO benchmark ~\cite{circo}. $^\dag$ indicates methods with multiple components (e.g., GPT-3.5, Qwen1.5-32B); we report \# parameters of components with known sizes. The \textcolor{gray}{CoCa-based MagicLens}$^{\ddagger}$ models are proprietary. Results in \textbf{bold} denote the best performances for each model scale.}
\vspace{-0.4cm}
\label{tab:CIRCO-full-results}
\end{table*}

\begin{table*}[h]
\begin{adjustbox}{width=\textwidth}
\begin{tabular}{l|cc|ccccccc}
\toprule
\multirow{2}{*}{\centering \textbf{Methods}} & \multirow{2}{*}{\centering \textbf{Backbone}} & \multirow{2}{*}{\centering \textbf{\# Params}} & \multicolumn{4}{c}{\textbf{Index Set}} & \multicolumn{3}{c}{\textbf{Subset Set}} \\ \cmidrule(lr){4-7} \cmidrule(lr){8-10} 
&           &           & R@1 & R@5 & R@10 & R@50 &R@1 &R@2 & R@3 \\\midrule
PALAVRA~\cite{cohen2022my} & CLIP-B &176M &16.6 &43.5 &58.5 &84.0 &41.6 &65.3 &80.9\\
PLI~\cite{PLI2023}           & BLIP-B            & 224M      & 27.2 &58.9 &71.4 &91.3 &55.1 &77.4 &89.1          \\
SEARLE~\cite{circo}        & CLIP-B            & 165M      &24.0 &53.4 &66.8 &89.8 &54.9 &76.6 &88.2 \\
CIReVL~\cite{CIReVL2024}  & CLIP-B            & 12.3B$^{\dag}$      & 23.9 &52.5 &66.0 &87.0 &60.2 &80.1 &90.2  \\
LDRE~\cite{ldre2024}          & CLIP-B            & 7.9B$^{\dag}$         & 25.7  & 55.1 &69.0 &89.9  &60.5 &80.7  &90.7   \\
MagicLens-B~\cite{magiclens}     & CLIP-B            & 166M      & 27.0 &58.0 &70.9 &91.1 &66.7 &83.9 &92.4      \\
\textcolor{gray}{MagicLens-B}$^{\ddagger}$~\cite{magiclens} & \textcolor{gray}{CoCa-B} & \textcolor{gray}{267M} & \textcolor{gray}{31.6} & \textcolor{gray}{64.0} & \textcolor{gray}{76.9} & \textcolor{gray}{93.8} & \textcolor{gray}{69.3} & \textcolor{gray}{86.0} & \textcolor{gray}{94.0} \\
\midrule
\rowcolor{gray!20} 
\textbf{MMRet-Base} & CLIP-B     & 149M      & \textbf{36.1}  & \textbf{68.1}  &  \textbf{79.5} & \textbf{94.7} & \textbf{71.6} & \textbf{87.2} & \textbf{94.0}        \\ 
\midrule
Pic2Word~\cite{saito2023pic2word}      & CLIP-L            & 429M      & 23.9 &51.7 &65.3 &87.8 &- &- &-       \\
PLI~\cite{PLI2023}           & CLIP-L            & 428M      & 25.5 &54.6 &67.6 &88.7 &55.6 &77.5 &89.5    \\
SEARLE~\cite{circo}        & CLIP-L            & 442M      & 24.2 &52.5 &66.3 &88.8 &53.8 &75.0 &88.2     \\
CIReVL~\cite{CIReVL2024}  & CLIP-L     & 12.5B$^{\dag}$    & 24.6 &52.3 &64.9 &86.3 &59.5 &79.9 &89.7  \\
LinCIR~\cite{lincir2024}        & CLIP-L            & 442M      & 25.0 &53.3 &66.7 &- &57.1 &77.4 &88.9        \\
CompoDiff~\cite{gu2023compodiff}     & CLIP-L            & 568M      & 18.2 &53.1 &70.8 &90.3 &57.4 &77.1 &87.9       \\
MagicLens-L~\cite{magiclens}     & CLIP-L            & 465M      & 30.1 &61.7 &74.4 &92.6 &68.1 &84.8 &93.2         \\
\textcolor{gray}{MagicLens-L}$^{\ddagger}$~\cite{magiclens} & \textcolor{gray}{CoCa-L} & \textcolor{gray}{613M} & \textcolor{gray}{33.3} & \textcolor{gray}{67.0} & \textcolor{gray}{77.9} & \textcolor{gray}{94.4} & \textcolor{gray}{70.9} & \textcolor{gray}{87.3} & \textcolor{gray}{94.5} \\
\midrule
\rowcolor{gray!20} 
\textbf{MMRet-Large} & CLIP-L     & 428M      & \textbf{38.0}  & \textbf{70.3}  & \textbf{81.1}   & \textbf{94.7}    & \textbf{73.2}  & \textbf{88.0} & \text{94.3}  \\ 
\midrule
Pic2Word~\cite{saito2023pic2word} & CLIP-H &987M &32.9 &63.1 &73.9 &- &62.2 &81.4 &91.2\\
SEARLE~\cite{circo} & CLIP-H &1.0B & 34.0 &64.0 &75.3 &- &64.6 &83.2 &92.8\\
LinCIR~\cite{lincir2024}    & CLIP-H     & 1.0B      & 33.8 &63.5 &73.4 &- &62.4 &81.5 &92.1\\
Pic2Word~\cite{saito2023pic2word} & CLIP-G &2.5B &30.4 &58.1 &69.2 &- &68.9 &85.5 &93.0\\
SEARLE~\cite{circo} & CLIP-G &2.6B & 34.8 &64.1 &75.1 &- &68.7 &84.7 &93.2\\
CompoDiff~\cite{gu2023compodiff} &CLIP-G &2.9B & 26.7 &55.1 &74.5 &92.0 &64.5 &82.4 &91.8 \\
CIReVL~\cite{CIReVL2024} &CLIP-G &14.6B$^{\dag}$ &34.7 &64.3 &75.1 &91.7 &68.0 &84.9 &93.2 \\
LinCIR~\cite{lincir2024}        & CLIP-G            & 2.6B      & 35.3 &64.7 &76.1 &- &63.4 &82.2 &92.0\\
LDRE~\cite{ldre2024}          & CLIP-G            & 10.3B$^{\dag}$       & 36.2  &66.4 &77.3 &94.0  &68.8 &85.7  &93.8   \\
IP-CIR~\cite{ip-cir2024}        & CLIP-G            & 43.8B$^{\dag}$         & 39.3  & 70.1 &80.0 &94.9  &70.0 &86.9 &94.2 \\
E5-V~\cite{e5-v2024}          & LLaVA-1.6         & 8.35B     & 33.9  & 64.1 &75.9 &93.5   &- & - & -\\
\midrule
\rowcolor{gray!20} 
\textbf{MMRet-MLLM} & LLaVA-1.6   & 7.57B     & \textbf{46.7}  & \textbf{76.0}  & \textbf{85.1}  & \textbf{96.5} &\textbf{75.4} &\textbf{89.6} &\textbf{95.7} \\ 
\bottomrule
\end{tabular}
\end{adjustbox}
\caption{Full results on the CIRR benchmark ~\cite{cirr-liu2021image}. $^\dag$ indicates methods with multiple components (e.g., GPT-3.5, Qwen1.5-32B); we report \# parameters of components with known sizes. The \textcolor{gray}{CoCa-based MagicLens}$^{\ddagger}$ models are proprietary. Results in \textbf{bold} denote the best performance for each model scale.}
\vspace{-0.4cm}
\vspace{-0.4cm}
\label{tab:CIRR-full-results}
\end{table*}

\begin{table*}[h]
\begin{adjustbox}{width=\textwidth}
\begin{tabular}{l|cc|cccccccc}
\toprule
\multirow{2}{*}{\centering \textbf{Methods}} & \multirow{2}{*}{\centering \textbf{Backbone}} & \multirow{2}{*}{\centering \textbf{\# Params}} & \multicolumn{2}{c}{\textbf{Dress}} & \multicolumn{2}{c}{\textbf{Shirt}} & \multicolumn{2}{c}{\textbf{Toptee}} & \multicolumn{2}{c}{\textbf{Overall}} \\ \cmidrule(lr){4-5} \cmidrule(lr){6-7} \cmidrule(lr){8-9} \cmidrule(lr){10-11}
&           &           & R@10 & R@50 & R@10 & R@50 &R@10 &R@50 & R@10 & R@50 \\\midrule
PALAVRA~\cite{cohen2022my} & CLIP-B &176M &17.3 &35.9 &21.5 &37.1 &20.6 &38.8 &19.8 &37.3\\
PLI~\cite{PLI2023}           & BLIP-B            & 224M      & 28.6 &50.8 &38.1 &57.8 &40.9 &62.7 &\textbf{35.9} &57.1          \\
SEARLE~\cite{circo}        & CLIP-B            & 165M      &18.5 &39.5 &24.4 &41.6 &25.7 &46.5 &22.9 &42.5 \\
CIReVL~\cite{CIReVL2024}  & CLIP-B            & 12.3B$^{\dag}$      & 25.3 &46.4 &28.4 &47.8 &31.2 &53.9 &28.3 &49.4 \\
LDRE~\cite{ldre2024}          & CLIP-B            & 7.9B$^{\dag}$         &27.4 &46.3 &20.0 &41.8 &27.1 &48.8  &24.8 &45.6  \\
MagicLens-B~\cite{magiclens}     & CLIP-B            & 166M      & 21.5 &41.3 &27.3 &48.8 &30.2 &52.3 &26.3 &47.4     \\
\textcolor{gray}{MagicLens-B}$^{\ddagger}$~\cite{magiclens} & \textcolor{gray}{CoCa-B} & \textcolor{gray}{267M} & \textcolor{gray}{29.0} & \textcolor{gray}{48.9} & \textcolor{gray}{36.5} & \textcolor{gray}{55.5} & \textcolor{gray}{40.2} & \textcolor{gray}{61.9} & \textcolor{gray}{35.2} & \textcolor{gray}{55.4} \\
\midrule
\rowcolor{gray!20} 
\textbf{MMRet-Base} & CLIP-B     & 149M      & 26.1  &  49.3 & 33.7  & 53.4 & 36.0 & 57.5 & 31.9   &  53.4    \\ 
\midrule
Pic2Word~\cite{saito2023pic2word}      & CLIP-L            & 429M      & 20.0 &40.2 &26.2 &43.6 &27.9 &47.4 &24.7 &43.7      \\
PLI~\cite{PLI2023}           & CLIP-L            & 428M      & 28.1 &51.1 &38.6 &58.5 &39.4 &62.7 &35.4 &57.4   \\
SEARLE~\cite{circo}        & CLIP-L            & 442M      & 20.5 &43.1 &26.9 &45.6 &29.3 &50.0 &25.6 &46.2     \\
CIReVL~\cite{CIReVL2024}  & CLIP-L     & 12.5B$^{\dag}$    & 24.8 &44.8 &29.5 &47.4 &31.4 &53.7 &28.6 &48.6  \\
LinCIR~\cite{lincir2024}        & CLIP-L            & 442M      & 20.9 &42.4 &29.1 &46.8 &28.8 &50.2 &26.3 &46.5      \\
CompoDiff~\cite{gu2023compodiff}     & CLIP-L            & 568M      &32.2 &46.3 &37.7 &49.1 &38.1 &50.6 &36.0 &48.6       \\
MagicLens-L~\cite{magiclens}     & CLIP-L            & 465M      & 25.5 &46.1 &32.7 &53.8 &34.0 &57.7 &30.7 &52.5         \\
\textcolor{gray}{MagicLens-L}$^{\ddagger}$~\cite{magiclens} & \textcolor{gray}{CoCa-L} & \textcolor{gray}{613M} & \textcolor{gray}{32.3} & \textcolor{gray}{52.7} & \textcolor{gray}{40.5} & \textcolor{gray}{59.2} & \textcolor{gray}{41.4} & \textcolor{gray}{63.0} & \textcolor{gray}{38.0} & \textcolor{gray}{58.2} \\
\midrule
\rowcolor{gray!20} 
\textbf{MMRet-Large} & CLIP-L     & 428M    & 29.7 & 50.3 &  37.0 & {56.1}   & {37.0}    & 59.3  & 34.6 &  55.2  \\ 
\midrule
Pic2Word~\cite{saito2023pic2word} & CLIP-H &987M &28.0 &51.5 &36.9 &56.0 &40.2 &62.0 &35.0 &56.5\\
SEARLE~\cite{circo} & CLIP-H &1.0B & 28.5 &51.1 &36.5 &55.5 &38.8 &60.9 &34.6 &55.8\\
LinCIR~\cite{lincir2024}    & CLIP-H     & 1.0B      & 29.8 &52.1 &36.9 &57.8 &42.1 &62.5 &36.3 &57.5\\
Pic2Word~\cite{saito2023pic2word} & CLIP-G &2.5B &25.4 & 47.7 &33.2 &50.4 &35.2 &57.6 &31.3 &51.9\\
SEARLE~\cite{circo} & CLIP-G &2.6B & 28.2 &50.3 &36.5 &55.4 &39.8 &61.5 &34.8 &55.7\\
CompoDiff~\cite{gu2023compodiff} &CLIP-G &2.9B & 37.8 &49.1 &41.3 &55.2 &44.3 &56.4 &39.0 &51.7 \\
CIReVL~\cite{CIReVL2024} &CLIP-G &14.6B$^{\dag}$ &27.1 &49.5 &33.7 &51.4 &35.8 &56.1 &32.2 &52.4 \\
LinCIR~\cite{lincir2024}        & CLIP-G            & 2.6B      & 38.1 &60.9 &46.8 &65.1 &50.5 &71.1 &45.1 &65.7\\
LDRE~\cite{ldre2024}          & CLIP-G            & 10.3B$^{\dag}$  &35.9 &58.6 &26.1 &51.1 &35.4 &56.7 &32.5 &55.5 \\
IP-CIR~\cite{ip-cir2024}        & CLIP-G            & 43.8B$^{\dag}$     &48.0 &66.7 &39.0 &61.0 &50.2 &71.1  &\textbf{45.7} &66.3 \\
E5-V~\cite{e5-v2024}          & LLaVA-1.6         & 8.35B     &36.4 &56.4 &23.8 &47.5 &35.3 &57.5 &31.8 &53.8  \\
\midrule
\rowcolor{gray!20} 
\textbf{MMRet-MLLM} & LLaVA-1.6   & 7.57B     &  29.1 & 50.5  & 38.5  & 58.6 & 39.2 & 60.6 & 35.6 & 56.6 \\ 
\bottomrule
\end{tabular}
\end{adjustbox}
\caption{Full results on the FashionIQ benchmark ~\cite{fashioniq-wu2021fashion}. $^\dag$ indicates methods with multiple components (e.g., GPT-3.5, Qwen1.5-32B); we report \# parameters of components with known sizes. The \textcolor{gray}{CoCa-based MagicLens}$^{\ddagger}$ models are proprietary. Results in \textbf{bold} denote the best performance for each model scale.}
\vspace{-0.4cm}
\vspace{-0.4cm}
\label{tab:FashionIQ-full-results}
\end{table*}

\begin{table*}[h]
\begin{adjustbox}{width=\textwidth}
\begin{tabular}{l|cc|ccccccccccccc}
\toprule
\multirow{2}{*}{\centering \textbf{Methods}} & \multirow{2}{*}{\centering \textbf{Backbone}} & \multirow{2}{*}{\centering \textbf{\# Params}} & \multicolumn{3}{c}{\textbf{Focus Attribute}} & \multicolumn{3}{c}{\textbf{Change Attribute}} & \multicolumn{3}{c}{\textbf{Focus Object}} & \multicolumn{3}{c}{\textbf{Change Object}} & \multicolumn{1}{c}{\textbf{Avg}}\\ \cmidrule(lr){4-6} \cmidrule(lr){7-9} \cmidrule(lr){10-12} \cmidrule(lr){13-15} \cmidrule(lr){16-16}
&           &           & R@1 & R@2 & R@3 & R@1 & R@2 & R@3 & R@1 & R@2 & R@3 & R@1 & R@2 & R@3 & R@1 \\\midrule
CIReVL~\cite{CIReVL2024}  & CLIP-B            & 12.3B$^{\dag}$      & 17.9 &29.4 &40.4 &14.8 &25.8 &35.8 &14.6 &24.3 &33.3 &16.1 &27.8 &37.6 &15.9 \\
MagicLens-B~\cite{magiclens}     & CLIP-B            & 166M      & 15.5 &28.4 &39.1 &12.3 &23.0 &32.1 &14.4 &26.2 &35.5 &17.7 &28.4 &39.2 &15.0     \\
\textcolor{gray}{MagicLens-B}$^{\ddagger}$~\cite{magiclens} & \textcolor{gray}{CoCa-B} & \textcolor{gray}{267M} & \textcolor{gray}{16.2} & \textcolor{gray}{27.8} & \textcolor{gray}{38.6} & \textcolor{gray}{16.2} & \textcolor{gray}{27.2} & \textcolor{gray}{36.6} & \textcolor{gray}{17.1} & \textcolor{gray}{27.7} & \textcolor{gray}{38.2} & \textcolor{gray}{20.2} & \textcolor{gray}{32.2} & \textcolor{gray}{42.9} & \textcolor{gray}{17.4}\\
\midrule
\rowcolor{gray!20} 
\textbf{MMRet-Base} & CLIP-B     & 149M      &18.3   & 30.9   & 39.6  &15.2  & 25.6 & 34.8 &16.6  & 27.3  & 35.8  &21.7 &34.9 &  45.0& \textbf{18.0} \\ 
\midrule
Pic2Word~\cite{saito2023pic2word}      & CLIP-L            & 429M      & 15.7 &28.2 &38.7 &13.9 &24.7 &33.1 &8.4 &18.0 &25.8 &6.7 &15.1 &24.0 &11.2      \\
SEARLE~\cite{circo}        & CLIP-L            & 442M      & 17.0 &29.7 &40.7 &16.4 &25.3 &34.1 &8.0 &16.9 &25.6 &7.9 &16.8 &24.8 &12.3     \\
CIReVL~\cite{CIReVL2024}  & CLIP-L     & 12.5B$^{\dag}$    & 19.5 &31.8 &42.0 &14.4 &26.0 &35.2 &12.3 &21.8 &30.5 &17.2 &28.9 &37.6 &15.9  \\
LinCIR~\cite{lincir2024}        & CLIP-L            & 442M      & 16.9 &30.0 &41.5 &16.2 &28.0 &36.8 &8.3 &17.4 &26.2 &7.4 &15.7 &25.0 &12.2      \\
CompoDiff~\cite{gu2023compodiff}     & CLIP-L            & 568M      &13.5 &24.3 &36.1 &19.2 &28.6 &37.2 &8.1 &16.4 &25.1 &18.7 &31.7 &40.6 &14.9       \\
MagicLens-L~\cite{magiclens}     & CLIP-L            & 465M      & 16.1 &28.2 &39.0 &15.6 &27.5 &36.3 &16.3 &26.2 &35.5 &17.1 &29.5 &39.7 &16.3        \\
\textcolor{gray}{MagicLens-L}$^{\ddagger}$~\cite{magiclens} & \textcolor{gray}{CoCa-L} & \textcolor{gray}{613M} & \textcolor{gray}{16.6} & \textcolor{gray}{28.7} & \textcolor{gray}{39.3} & \textcolor{gray}{16.0} & \textcolor{gray}{27.5} & \textcolor{gray}{36.5} & \textcolor{gray}{15.7} & \textcolor{gray}{27.6} & \textcolor{gray}{37.3} & \textcolor{gray}{18.7} & \textcolor{gray}{31.7} & \textcolor{gray}{40.2} & \textcolor{gray}{16.7}\\
\midrule
\rowcolor{gray!20} 
\textbf{MMRet-Large} & CLIP-L     & 428M    &18.4 & 30.0 & 38.5  & 15.4   & 27.6    & 35.7  &17.4  & 26.6 & 36.3 & 21.0 & 34.0 & 42.4 &\textbf{18.1}  \\ 
\midrule
Pic2Word~\cite{saito2023pic2word} & CLIP-H &987M &18.6 &30.7 &42.1 &13.2 &23.9 &33.1 &9.2 &17.6 &27.1 &6.6 &16.5 &25.4 &11.9\\
SEARLE~\cite{circo} & CLIP-H &1.0B & 18.8 &31.5 &42.3 &15.5 &26.9 &35.9 &10.6 &18.7 &26.5 &8.5 &17.9 &26.2 &13.3\\
LinCIR~\cite{lincir2024}    & CLIP-H     & 1.0B      & 19.6 &31.5 &41.6 &16.6 &27.6 &37.5 &9.8 &18.8 &27.9 &9.0 &17.6 &25.7 &13.8\\
Pic2Word~\cite{saito2023pic2word} & CLIP-G &2.5B &12.5 &23.4 &33.7 &11.7 &21.9 &30.9 &9.9 &19.3 &27.4 &8.6 &18.2 &26.1 &10.7\\
SEARLE~\cite{circo} & CLIP-G &2.6B & 16.3 &29.4 &40.7 &16.2 &27.3 &35.5 &10.8 &18.2 &27.9 &8.3 &15.6 &25.8 &12.9\\
CompoDiff~\cite{gu2023compodiff} &CLIP-G &2.9B & 14.3 &26.7 &38.4 &19.7 &28.8 &37.4 &9.2 &19.1 &25.8 &18.7 &31.7 &40.2 &15.5 \\
CIReVL~\cite{CIReVL2024} &CLIP-G &14.6B$^{\dag}$ &20.5 &34.0 &44.5 &16.1 &28.6 &39.4 &14.7 &25.2 &33.0 &18.1 &31.2 &41.0 &17.4 \\
LinCIR~\cite{lincir2024}        & CLIP-G            & 2.6B      & 19.1 &33.0 &42.3 &17.6 &30.2 &38.1 &10.1 &19.1 &28.1 &7.9 &16.3 &25.7 &13.7\\
\midrule
\rowcolor{gray!20} 
\textbf{MMRet-MLLM} & LLaVA-1.6   & 7.57B     &18.4   &31.4   &41.0   &16.7  &27.7 &36.4 &22.4 &32.5  &41.6 &26.9 &40.4 &49.9 &\textbf{21.1}\\ 
\bottomrule
\end{tabular}
\end{adjustbox}
\caption{Full results on the GeneCIS benchmark ~\cite{genecis2023}. $^\dag$ indicates methods with multiple components (e.g., GPT-3.5, Qwen1.5-32B); we report \# parameters of components with known sizes. The \textcolor{gray}{CoCa-based MagicLens}$^{\ddagger}$ models are proprietary. Results in \textbf{bold} denote the best performance for each model scale.}
\vspace{-0.4cm}
\vspace{-0.4cm}
\label{tab:GeneCIS-full-results}
\end{table*}

\begin{table*}[ht]
\centering
\renewcommand{\arraystretch}{1.2}
\setlength{\tabcolsep}{4pt}
\resizebox{\textwidth}{!}{
\begin{tabular}{lcccccccccccc}
\toprule
\multirow{2}{*}{\textbf{Task}} & \multicolumn{8}{c}{\textbf{Zero-shot}} & \multicolumn{2}{c}{\textbf{Fine-Tune}}\\ \cmidrule(lr){2-9} \cmidrule(lr){10-11} 
 & \textbf{CLIP} & \textbf{OpenCLIP} & \textbf{SigLIP} & \textbf{BLIP2} & \textbf{MagicLens} & \textbf{E5-V} & \textbf{UniIR}  & \textbf{MMRet} & \textbf{VLM2Vec} & \textbf{MMRet}\\ 
\midrule
\rowcolor[HTML]{FAE3D6}
\multicolumn{11}{l}{\textbf{Classification (10 tasks)}} \\
ImageNet-1K & 55.8 & 63.5 & 45.4 & 10.3 & 48.0 & 9.6 & 53.7 & 49.1 & 65.6 & 58.8\\
N24News & 34.7 & 38.6 & 13.9 & 36.0 & 33.7 & 23.4 & 33.9 & 45.8 & 79.5 & 71.3\\
HatefulMemes & 51.1 & 51.7 & 47.2 & 49.6 & 49.0 & 49.7 & 51.0 & 51.0 &67.1 & 53.7\\
VOC2007 & 50.7 & 52.4 & 64.3 & 52.1 & 51.6 & 49.9 & 62.7 & 74.6 & 88.6 & 85.0\\
SUN397 & 43.4 & 68.8 & 39.6 & 34.5 & 57.0 & 33.1 & 61.7 & 60.1 & 72.7 & 70.0\\
\rowcolor[HTML]{E8E8E8} Place365 & 28.5 & 37.8 & 20.0 & 21.5 & 31.5 & 8.6 & 38.0 & 35.3 & 42.6 & 43.0\\
\rowcolor[HTML]{E8E8E8} ImageNet-A & 25.5 &14.2 &42.6 &3.2 &8.0 &2.0 &12.9 & 31.6 &19.3 & 36.1\\
\rowcolor[HTML]{E8E8E8} ImageNet-R & 75.6 & 83.0 & 75.0 & 39.7 & 70.9 & 30.8 & 61.6 & 66.2& 70.2 & 71.6\\
\rowcolor[HTML]{E8E8E8} ObjectNet & 43.4 & 51.4 & 40.3 & 20.6 & 31.6 & 7.5 & 37.1 & 49.2 & 29.5 & 55.8\\
\rowcolor[HTML]{E8E8E8} Country-211 & 19.2 & 16.8 & 14.2 & 2.5 & 6.2 & 3.1 & 8.8 & 9.3 & 13.0 & 14.7\\
\textit{All Classification} & 42.8 & 47.8 & 40.3 & 27.0 & 38.8 & 21.8 & 42.1 & 47.2 & 54.8 & 56.0\\
\midrule
\rowcolor[HTML]{F2F2FA}
\multicolumn{11}{l}{\textbf{VQA (10 tasks)}} \\
OK-VQA & 7.5 & 11.5 & 2.4 & 8.7 & 12.7 & 8.9 & 24.5 & 28.0 & 63.2 & 73.3\\
A-OKVQA & 3.8 & 3.3 & 1.5 & 3.2 & 2.9 & 5.9 & 10.6 & 11.6 & 50.2 & 56.7\\
DocVQA & 4.0 & 5.3 & 4.2 & 2.6 & 3.0 & 1.7 & 5.6 & 12.6 & 78.4 & 78.5\\
InfographicsVQA & 4.6 & 4.6 & 2.7 & 2.0 & 5.9 & 2.3 & 5.0 & 10.6 & 40.8 & 39.3\\
ChartQA & 1.4 & 1.5 & 3.0 & 0.5 & 0.9 & 2.4 & 1.8 & 2.4 & 59.0 & 41.7\\
Visual7W & 4.0 &2.6 &1.2 &1.3 &2.5 &5.8 &12.3 & 9.0 & 47.7& 49.5\\
\rowcolor[HTML]{E8E8E8} ScienceQA & 9.4 & 10.2 & 7.9 & 6.8 & 5.2 & 3.6 & 11.6 & 23.3 & 43.4 & 45.2\\
\rowcolor[HTML]{E8E8E8} VizWiz & 8.2 & 6.6 & 2.3 & 4.0 & 1.7 & 2.6 & 19.2 & 25.9 & 39.2 & 51.7\\
\rowcolor[HTML]{E8E8E8} GQA & 41.3 & 52.5 & 57.5 & 9.7 & 43.5 & 7.8 & 49.3 & 41.3 & 60.7 & 59.0\\
\rowcolor[HTML]{E8E8E8} TextVQA & 7.0 & 10.9 & 1.0 & 3.3 & 4.6 & 3.2 & 10.6 & 18.9 & 66.1 & 79.0\\
\textit{All VQA} & 9.1 & 10.9 & 8.4 & 4.2 & 8.3 & 4.9 & 15.0 & 18.4 & 54.9 & 57.4\\
\midrule
\rowcolor[HTML]{DFF2DF}
\multicolumn{11}{l}{\textbf{Retrieval (12 tasks)}} \\
VisDial & 30.7 & 25.4 & 21.5 & 18.0 & 24.8 & 9.2 & 37.6 & 62.6 & 73.3 & 83.0\\
CIRR & 12.6 & 15.4 & 15.1 & 9.8 & 39.1 & 6.1 & 53.2 & 65.7 & 47.8 &  61.4\\
VisualNews\_t2i & 78.9 & 74.0 & 51.0 & 48.1 & 50.7 & 13.5 & 63.6 & 45.7 & 67.2 & 74.2\\
VisualNews\_i2t & 79.6 & 78.0 & 52.4 & 13.5 & 21.1 & 8.1 & 68.8 & 53.4 & 70.7 & 78.1\\
MSCOCO\_t2i & 59.5 & 63.6 & 58.3 & 53.7 & 54.1 & 20.7 & 72.0 & 68.7 & 70.6 & 78.6\\
MSCOCO\_i2t & 57.7 &62.1 &55.0 &20.3 &40.0 &14.0 &74.1 & 56.7 &66.5 & 72.4\\
NIGHTS & 60.4 & 66.1 & 62.9 & 56.5 & 58.1 & 4.2 & 69.7 & 59.4& 66.1 & 68.3\\
WebQA & 67.5 & 62.1 & 58.1 & 55.4 & 43.0 & 17.7 & 86.3 & 76.3&88.1 & 90.2 \\
\rowcolor[HTML]{E8E8E8}FashionIQ & 11.4 & 13.8 & 20.1 & 9.3 & 11.2 & 2.8 & 39.3 & 31.5& 12.9 & 54.9\\
\rowcolor[HTML]{E8E8E8}Wiki-SS-NQ & 55.0 &44.6 &55.1 &28.7 &18.7 &8.6 &11.3 & 25.4&56.6 & 24.9\\
\rowcolor[HTML]{E8E8E8}OVEN &  41.1 &45.0 &56.0 &39.5 &1.6 &5.9 &66.6 & 73.0 &47.3 & 87.5\\
\rowcolor[HTML]{E8E8E8}EDIS & 81.0 & 77.5 & 23.6 & 54.4 & 62.6 & 26.8 & 78.2 & 59.9 &79.9 & 65.6\\
\textit{All Retrieval} & 53.0 & 52.3 & 31.6 & 33.9 & 35.4 & 11.5 & 60.1 & 56.5 &62.3 & 69.9\\
\midrule
\rowcolor[HTML]{FCE4D6}
\multicolumn{11}{l}{\textbf{Visual Grounding (4 tasks)}} \\
MSCOCO & 33.8 & 34.5 & 46.4 & 28.9 & 22.1 & 10.8 & 46.6 & 42.7 &67.3 & 76.8\\
\rowcolor[HTML]{E8E8E8}RefCOCO & 56.9 & 54.2 & 70.8 & 47.4 & 22.8 & 11.9 & 67.8 & 69.3 &84.7& 89.8\\
\rowcolor[HTML]{E8E8E8}RefCOCO-matching & 61.3 &68.3 &50.8 &59.5 &35.6 &38.9 &62.9 & 63.2 &79.2 & 90.6\\
\rowcolor[HTML]{E8E8E8}Visual7W-pointing & 55.1 & 56.3 & 70.1 & 52.0 & 23.4 & 14.3 & 71.3 & 73.5 &86.8 & 77.0\\
\textit{All Visual Grounding} & 51.8 & 53.3 & 59.5 & 47.0 & 26.0 & 19.0 & 62.2 &62.2 &79.5 & 83.6\\
\midrule
\rowcolor[HTML]{E6F7FF}
\multicolumn{11}{l}{\textbf{Final Score (36 tasks)}} \\
\text{All} & 37.8 & 39.7 & 34.8 & 25.2 & 27.8 & 13.3 & 42.8 & 44.0 & 60.1 & 64.1\\
\text{All IND} & 37.1 & 39.3 & 32.3 & 25.3 & 31.0 & 14.9 & 44.7 & 43.5 & 66.5 & 59.1\\
\text{All OOD} & 38.7 & 40.2 & 38.0 & 25.1 & 23.7 & 11.5 & 40.4 & 44.3 & 52.0 & 68.0\\
\bottomrule
\end{tabular}
}
\caption{The detailed results on the MMEB benchmark ~\cite{vlm2vec2024}. We report the performance of our MMRet under both zero-shot and fine-tuning settings.}

\label{tab:mmeb_full_performance}
\end{table*}

\end{document}